\documentclass[journal]{IEEEtran}
\usepackage{amsmath,amsfonts,amssymb}
\usepackage{graphicx}
\usepackage{cite}
\usepackage{booktabs}
\usepackage{algorithm}
\usepackage{algorithmic}
\usepackage{url}
\usepackage{multirow}
\usepackage{color}
\usepackage{subcaption}

\usepackage{hyperref}
\hypersetup{
    colorlinks=true,
    linkcolor=blue,
    citecolor=blue,
    urlcolor=blue,
}

\begin{document}


\title{PhyG-MoE: A Physics-Guided Mixture-of-Experts Framework for Energy-Efficient GNSS Interference Recognition}

\author{Zhihan Zeng, \IEEEmembership{Graduate Student Member, IEEE}, Yang Zhao, \IEEEmembership{Member, IEEE},\\
Kaihe Wang, \IEEEmembership{Graduate Student Member, IEEE}, Dusit Niyato, \IEEEmembership{Fellow, IEEE},\\
Yue Xiu, \IEEEmembership{Member, IEEE}, 
Lu Chen, Zhongpei Zhang, Ning Wei, \IEEEmembership{Member, IEEE}
\thanks{Zhihan Zeng, Yue Xiu, Zhongpei Zhang and Ning Wei are with the National Key Laboratory of Wireless Communications, University of Electronic Science and Technology of China (UESTC), Chengdu 611731, China (E-mail: 202511220608@std.uestc.edu.cn, xiuyue12345678@163.com, zhangzp@uestc.edu.cn, wn@uestc.edu.cn). Kaihe Wang is with the University of Electronic Science and Technology of China (UESTC), Chengdu 611731, China (E-mail: khewang@yeah.net). Y. Zhao is with the Nanyang Technological University, Singapore 639798 (e-mail: zhao0466@e.ntu.edu.sg). D. Niyato is with the College of Computing and Data Science, Nanyang Technological University, Singapore 639798 (e-mail: DNIYATO@ntu.edu.sg). Lu Chen is with the Anhui Science and Technology University Chuzhou 239000, China and Nanjing University Of Information Science $\&$ technology, Nanjing 210044, China (E-mail: chenlu@ahstu.edu.cn). The corresponding author is Lu Chen.}}
\maketitle

\begin{abstract} 
Complex electromagnetic interference increasingly compromises Global Navigation Satellite Systems (GNSS), threatening the reliability of Space-Air-Ground Integrated Networks (SAGIN). Although deep learning has advanced interference recognition, current static models suffer from a \textbf{fundamental limitation}: they impose a fixed computational topology regardless of the input's physical entropy. This rigidity leads to severe resource mismatch, where simple primitives consume the same processing cost as chaotic, saturated mixtures. To resolve this, this paper introduces PhyG-MoE (Physics-Guided Mixture-of-Experts), a framework designed to \textbf{dynamically align model capacity with signal complexity}. Unlike static architectures, the proposed system employs a spectrum-based gating mechanism that routes signals based on their spectral feature entanglement. A high-capacity TransNeXt expert is activated on-demand to disentangle complex features in saturated scenarios, while lightweight experts handle fundamental signals to minimize latency. Evaluations on 21 jamming categories demonstrate that PhyG-MoE achieves an overall accuracy of 97.58\%. By resolving the intrinsic conflict between static computing and dynamic electromagnetic environments, the proposed framework significantly reduces computational overhead without performance degradation, offering a viable solution for resource-constrained cognitive receivers. 
\end{abstract}


\section{Introduction}
\label{sec:introduction}

\IEEEPARstart{W}{ith} the continuous evolution of next-generation communication technologies \cite{liang2025generative, tang2026digital}, the Space-Air-Ground Integrated Network (SAGIN) has emerged as a fundamental architecture designed to provide seamless and ubiquitous connectivity across terrestrial, aerial, and space segments \cite{saad2020vision, nguyen20226g, wang2026cluster, zhang2026joint}. Within this multi-dimensional framework, the Global Navigation Satellite System (GNSS) functions as the core infrastructure underpinning network synchronization and operational coordination. Beyond providing essential Positioning, Navigation, and Timing (PNT) services, GNSS is increasingly critical for supporting high-precision network synchronization, autonomous Uncrewed Aerial Vehicle (UAV) guidance \cite{wang2026uav}, and beam alignment in Integrated Sensing and Communication (ISAC) systems \cite{liu2026star, kato2019optimizing, chen2024cognitive}. Consequently, the reliability and resilience of GNSS signals directly determine the stability of the entire integrated network ecosystem.

However, the GNSS downlink faces increasingly severe challenges in the physical layer due to the inherent vulnerability of the open wireless channel. Navigation signals, transmitted from Medium Earth Orbit (MEO) satellites at altitudes exceeding 20,000 km, undergo substantial free-space path loss before reaching the Earth's surface. Coupled with strict limitations on satellite transmission power and the transparent structure of civil spreading codes, the received signal power at terrestrial terminals is typically extremely low—often submerged below the thermal noise floor \cite{gao2016protecting}. This inherent power deficit renders GNSS receivers highly sensitive to intentional Radio Frequency Interference (RFI). The threat landscape has further deteriorated with the widespread availability of programmable Software Defined Radios (SDRs), which facilitate the synthesis of sophisticated compound interference. Adversaries can now mask sweep jammers with wideband noise or superimpose multiple waveforms to create complex spectral signatures \cite{liu2024gnss}. Unlike simple single-source primitives, these compound signals exhibit highly entangled Time-Frequency (TF) features, significantly degrading the performance of conventional anti-jamming processors and threatening the integrity of critical network services \cite{xiao2025compound,11353414,11355857,11316665,11098592,11220909,11207524,11346858,11316633,11108293,10797657}.

Accurate identification of the interference type is a prerequisite for applying specific suppression techniques, such as adaptive notch filtering or frequency-domain excision \cite{wesson2018gnss, gamba2019performance}. Conventional methods relying on statistical hypothesis testing are effective for jamming primitives but struggle to distinguish complex signals with overlapping spectral characteristics \cite{wang2017statistical, wang2018detection}. Deep Learning (DL) methods have recently demonstrated capabilities in learning discriminative features from TF distributions, outperforming manual feature engineering \cite{mehr2025deep, cai2019waterfall}.

A practical engineering issue remains in existing DL-based solutions: static computational cost. Most models execute a fixed number of Floating-Point Operations (FLOPs) per sample, ignoring signal complexity \cite{jia2025lightweight}. For SWaP-constrained satellite receivers, expending high computational resources to identify a simple single-tone jammer is inefficient \cite{chen2024cognitive, vandermerwe2024optimal}. This redundancy increases power consumption and processing latency, limiting the deployment of advanced models in energy-sensitive scenarios.

To optimize the trade-off between recognition accuracy and efficiency, we propose PhyG-MoE, a physics-guided Mixture-of-Experts (MoE) framework. The system implements a dynamic computing strategy based on the signal's Power Spectral Density (PSD). High-capacity experts activate only for complex compound signals to disentangle coupled features, while lightweight experts handle fundamental primitives. This on-demand allocation reduces energy consumption without compromising detection performance.

The main contributions of this paper are summarized as follows:

\begin{itemize}
    \item We synthesize a hierarchical jamming dataset covering 21 categories, ranging from primitives to saturated triple-component mixtures. This dataset explicitly models spectral entanglement and power variations in compound jamming, providing a realistic benchmark for cognitive recognition.
    
    \item We propose the PhyG-MoE framework, integrating a Spectrum-Aware Gating Mechanism with a heterogeneous expert pool. This design mitigates the computational redundancy of static models by leveraging physical spectral features to dynamically allocate resources.
    
   \item We design distinct experts tailored to specific physical interference attributes: a Co-TransNeXt expert incorporates Coordinate Attention Gated Linear Units (CoordAttGLU) to explicitly model long-range directional dependencies in entangled TF spectrograms, while an SK-GhostNet expert leverages Selective Kernels (SK) to adaptively adjust receptive fields, thereby accommodating the drastic bandwidth variations between narrowband and wideband jamming.
    
    \item Simulation results demonstrate that PhyG-MoE achieves an overall accuracy of 97.58\%. The dynamic routing strategy minimizes average FLOPs compared to static dual-stream baselines, balancing accuracy and efficiency.
\end{itemize}

The remainder of this paper is organized as follows. Section \ref{sec:related_work} reviews related works. Section \ref{sec:system_model} details the signal model. Section \ref{sec:signal_analysis} analyzes TF and spectral features. Section \ref{sec:methodology} elaborates on the MoE framework. Section \ref{sec:simulation} presents performance analysis, followed by the conclusion in Section \ref{sec:conclusion}.

\section{Related Work}
\label{sec:related_work}

This section reviews the evolution of interference classification and mitigation techniques, transitioning from conventional signal processing approaches to data-driven deep learning models, and finally discussing the emerging paradigm of dynamic neural networks.

\subsection{Conventional Signal Processing Based Approaches}
Traditional GNSS receiver protection mechanisms rely heavily on signal processing techniques that exploit the physical disparities between authentic satellite signals and interference \cite{gao2016protecting, wesson2018gnss}. Early methodologies primarily focused on statistical anomaly detection in the time domain or spectral analysis in the frequency domain. Techniques such as Adaptive Notch Filters (ANF) and Frequency Locked Loops (FLL) have been widely adopted for tracking and excising Narrowband Interference (NBI) and simple sweep jamming \cite{gamba2019performance, borio2014multistate, qin2020assessment}. Similarly, frequency-domain excision methods utilizing the Fast Fourier Transform (FFT) or sub-band Automatic Gain Control (AGC) are effective in mitigating Continuous Wave Interference (CWI) and partial-band noise \cite{zhang2011effect, garzia2021subband}.

To address non-stationary signals, such as chirp jammers, TF analysis tools have been extensively investigated. The Short-Time Fourier Transform (STFT) and Wigner-Ville Distribution (WVD) are commonly employed to visualize spectral variations \cite{wang2018detection, wang2017statistical}. Advanced transforms, including the Fractional Fourier Transform (FrFT) and Zak transform, offer superior energy concentration properties for chirp signals, facilitating precise parameter estimation \cite{sun2024frft, sun2024novel, alvarez2025chirp, luo2024zak}. Additionally, wavelet transforms provide multi-resolution analysis capabilities for isolating interference coefficients \cite{dovis2011wavelet}. To differentiate interference from system noise, statistical hypothesis testing methods are frequently utilized as pre-correlation detectors \cite{wang2017statistical, wang2018detection}.

However, conventional approaches face two primary limitations. First, they rely significantly on manual threshold settings and prior knowledge of the interference model, which lacks robustness under dynamic channel conditions \cite{wesson2018gnss}. Second, they encounter difficulties in distinguishing complex overlapping interferences, such as mixed sweep and pulse signals, where statistical features become ambiguous or the computational cost of optimal filtering becomes prohibitive for real-time receivers \cite{dasilva2023nmf, li2025multipolarization}.

\subsection{Deep Learning Based Interference Classification}
To mitigate the limitations of handcrafted features, DL has been introduced to automatically learn discriminative representations from raw data or TF spectrograms \cite{saad2020vision, nguyen20226g}. Convolutional Neural Networks (CNNs) have demonstrated significant efficacy in classifying jamming patterns by treating TF spectrograms as visual inputs \cite{liu2019deep, ferre2019jammer, cai2019waterfall}. Recent advancements have extended this paradigm by employing YOLO-based object detection frameworks to simultaneously classify and localize interference in the TF domain \cite{liu2024gnss}. To enhance classification performance under low Jamming-to-Noise Ratio (JNR) conditions, advanced architectures such as Deep Neural Networks (DNN) with fingerprint spectrums \cite{chen2022fingerprint}, multi-scale feature fusion networks \cite{jia2025lightweight}, and hierarchical classifiers \cite{jia2025lowpower, mehr2025deep} have been proposed.

Beyond standard CNNs, complex topologies have been explored to capture global dependencies. Graph Convolutional Networks (GCN) with adaptive weight learning have been applied to model non-Euclidean correlations among interference features \cite{li2025dualGCN}. For compound interference recognition, dual-stream networks integrating TF images and power spectrum features have outperformed single-stream models \cite{xiao2025compound}. Furthermore, segmentation-based approaches like TF-Unet have been utilized for pixel-level interference mitigation \cite{song2025tfunet, song2025sensing}. In the context of SAGIN, AI-driven optimization has proven essential for managing complex traffic and interference dynamics \cite{kato2019optimizing, liu2018sagin}. Research has also explored lightweight alternatives, such as GRU-based models \cite{mehr2025deep, mehr2025gru} or combining machine learning with efficient signal processing pre-processing to reduce the computational load on edge devices \cite{vandermerwe2024optimal, jia2025lightweight}.

Despite these advancements, a critical gap remains: the majority of existing DL models employ static architectures. Regardless of whether the input is a simple single-tone jammer or a highly complex mixed interference, the network executes an identical, computationally intensive graph. This one-size-fits-all strategy results in redundant energy consumption and latency, which is suboptimal for resource-constrained satellite or receiver platforms \cite{chen2024cognitive}.

\subsection{Dynamic Neural Networks and Efficient Computing}
To address the efficiency bottleneck, dynamic neural networks, particularly the MoE paradigm, have emerged as a promising solution. MoE enables conditional computation by activating only a subset of network parameters based on the input features \cite{lepikhin2020gshard}. This approach has achieved success in natural language processing and computer vision tasks, such as large-scale forecasting \cite{xu2023slmoe} and unified image restoration frameworks \cite{an2025moe}.

Parallel to dynamic routing, lightweight architecture design has evolved significantly. The MobileNet series and GhostNet have demonstrated that efficient operations, such as depthwise separable convolutions, can achieve high performance with minimal FLOPs \cite{howard2019mobilenetv3, qin2024mobilenetv4, han2020ghostnet}. Furthermore, adaptive feature extraction mechanisms, such as SK networks, allow models to dynamically adjust their receptive fields \cite{li2019selective, li2023largesk, yang2023complexsk}. Recently, TransNeXt introduced biological-inspired foveal perception to Vision Transformers, effectively balancing global attention with local detail \cite{shi2024transnext}, while Tokens-to-Token (T2T) ViT optimizes feature tokenization for enhanced structure modeling \cite{yuan2021t2t}. Additionally, Hypergraph Convolutional Networks have been proposed to handle multi-modal correlations efficiently \cite{nong2023adaptive}.

Although MoE and lightweight designs are widely adopted in general computer vision, a spectrum-aware dynamic routing mechanism specifically designed for the drastic complexity variations in GNSS interference remains unexplored. Existing cognitive radio methods often rely on simple switching logic rather than deep feature-based routing \cite{chen2024cognitive}. To the best of our knowledge, no prior work has successfully integrated PSD-guided dynamic routing with a heterogeneous expert pool to achieve an optimal trade-off between recognition accuracy and computational efficiency in GNSS interference classification.

\begin{figure}[htbp]
    \centering
    \includegraphics[width=0.9\linewidth]{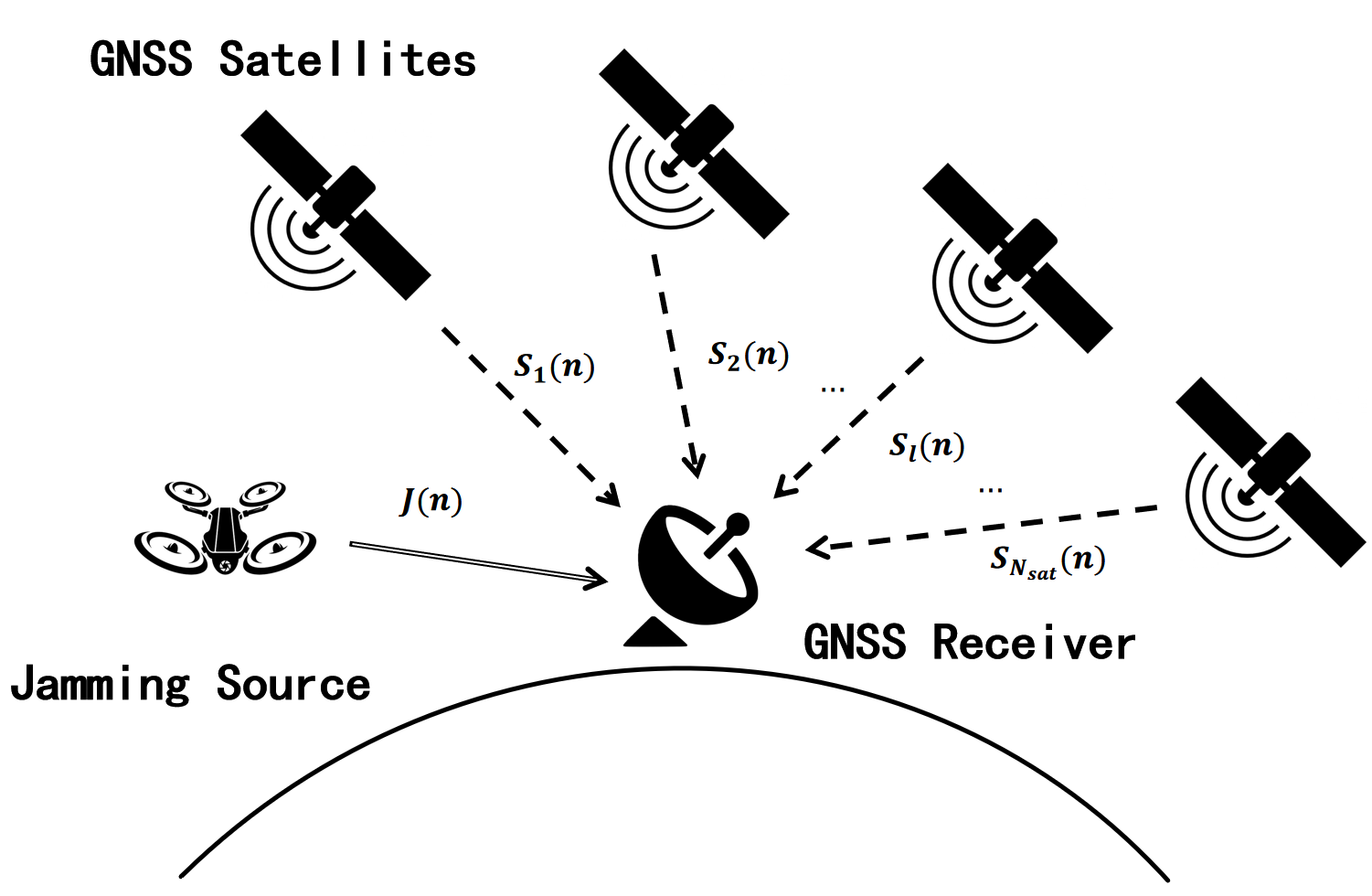}
    \caption{Illustration of the GNSS interference scenario. The receiver captures the superposition of authentic satellite signals, jamming signals, and noise.}
    \label{fig:gnss_model}
\end{figure}

\section{System Model}
\label{sec:system_model}

We frame the interference recognition task as the critical spectrum sensing stage of a cognitive GNSS receiver, performed prior to anti-jamming mitigation \cite{gao2016protecting, chen2024cognitive}. In this framework, the receiver monitors the electromagnetic spectrum to detect and classify jamming signals, enabling adaptive switching of suppression algorithms \cite{mehr2025deep, ferre2019jammer}. This section formulates the signal reception model and defines the mathematical structure of the jamming dataset, encompassing both fundamental primitives and compound interference architectures \cite{xiao2025compound}.

\subsection{Cognitive Signal Reception Formulation}
The RF signal is captured by the antenna, down-converted, and digitized, as illustrated in Fig. \ref{fig:gnss_model}. Let $F_s$ denote the sampling frequency and $N$ represent the number of samples in a coherent observation window. The discrete-time complex baseband signal $r[n]$ is modeled as \cite{wang2018detection}:
\begin{equation}
    r[n] = \sum_{l=1}^{N_{sat}} \sqrt{P_s} s_l[n] + \mathcal{J}[n] + w[n],
    \label{eq:rx_signal}
\end{equation}
where $N_{sat}$ is the number of visible satellites, $P_s$ is the received power of the authentic signal, and $s_l[n]$ denotes the spreading code of the $l$-th satellite. The term $w[n] \sim \mathcal{CN}(0, \sigma_n^2)$ represents Additive White Gaussian Noise (AWGN), and $\mathcal{J}[n]$ is the jamming component.

In GNSS contexts, the authentic signal arrives at the receiver surface with power levels well below the thermal noise floor (i.e., $P_s \ll \sigma_n^2$). Conversely, the jamming power $P_J$ is typically several orders of magnitude higher than both the signal and noise ($P_J \gg P_s$) \cite{wesson2018gnss}. Therefore, from the perspective of the interference classifier, the authentic signal $s[n]$ is statistically submerged in the background noise. We define the effective noise term $\tilde{w}[n] = w[n] + \sum \sqrt{P_s}s_l[n] \approx w[n]$. Consequently, the input to the cognitive perception module, denoted as $x[n]$, is modeled as:
\begin{equation}
    x[n] = \mathcal{J}[n] + \tilde{w}[n].
\end{equation}
The severity of the interference is quantified by the JNR, defined over the observation window as:
\begin{equation}
    \text{JNR}_{\text{dB}} = 10 \log_{10}\left( \frac{P_J}{\sigma_n^2} \right) = 10 \log_{10}\left( \frac{\frac{1}{N}\sum_{n=0}^{N-1} |\mathcal{J}[n]|^2}{\sigma_n^2} \right).
\end{equation}

\subsection{Fundamental Jamming Primitives}
To construct a comprehensive library, we first define the set of fundamental jamming primitives $\Omega = \{ \text{STJ, MTJ, LFM, Pulse, PBNJ} \}$ \cite{ferre2019jammer, vandermerwe2024optimal}. Let $t = n/F_s$ be the discrete time variable.

\subsubsection{Single-Tone Jamming (STJ)}
STJ concentrates energy at a specific frequency point. It is modeled as:
\begin{equation}
    \mathcal{J}_{\text{STJ}}[n] = e^{j(2\pi f_c t + \phi)},
\end{equation}
where the carrier frequency $f_c$ is modeled as a random variable uniformly distributed within the receiver bandwidth, i.e., $f_c \sim \mathcal{U}(-F_s/2, F_s/2)$, and $\phi \sim \mathcal{U}[0, 2\pi]$ is a random initial phase.

\subsubsection{Multi-Tone Jamming (MTJ)}
MTJ synthesizes a superposition of $K$ distinct sinusoidal waves to disrupt multiple sub-bands. The model is given by
\begin{equation}
\mathcal{J}_{\text{MTJ}}[n] = \sum_{k=1}^{K} e^{j(2\pi f_k t + \phi_k)},
\end{equation}

where $\{f_k\}_{k=1}^K$ are randomly selected frequencies to avoid fixed comb patterns \cite{ferre2019jammer} and $\phi_k \sim \mathcal{U}[0, 2\pi]$ denotes the random initial phase of the $k$-th tone, creating a challenging scenario for adaptive notch filters \cite{gamba2019performance, borio2014multistate, qin2020assessment}.

\subsubsection{Linear Frequency Modulation (LFM)}
LFM (chirp jamming) sweeps its instantaneous frequency linearly across a bandwidth $B_{sw}$ over a period $T_{sw}$. The signal is defined as \cite{alvarez2025chirp}:
\begin{equation}
    \mathcal{J}_{\text{LFM}}[n] = e^{j 2\pi (f_{start} t + \frac{1}{2} \mu t^2)},
\end{equation}
where $f_{start}$ denotes the initial frequency and $\mu = \pm B_{sw}/T_{sw}$ is the chirp rate. The parameters $f_{start}$ and $\mu$ are randomized to vary the sweep coverage and direction \cite{sun2024frft}.

\subsubsection{Pulse Jamming}
Pulse jamming is a non-stationary interference modulated by a binary masking sequence $w_{pulse}[n]$ given as:
\begin{equation}
    \mathcal{J}_{\text{Pulse}}[n] = w_{pulse}[n] \cdot e^{j 2\pi f_c t}.
\end{equation}
The gating function $w_{pulse}[n]$ is determined by the Pulse Repetition Interval (PRI) and pulse width $\tau$. Let $N_{PRI} = \lfloor \text{PRI} \cdot F_s \rfloor$ and $N_{\tau} = \lfloor \tau \cdot F_s \rfloor$ denote the corresponding number of samples. The gating function is given as follows:
\begin{equation}
    w_{pulse}[n] = \begin{cases} 
    1, & \text{if } n \pmod{N_{PRI}} < N_{\tau} \\
    0, & \text{otherwise}
    \end{cases},
\end{equation}
where the duty cycle $D = N_{\tau}/N_{PRI}$ characterizes the temporal density of the interference \cite{garzia2021subband}.

\subsubsection{Partial-Band Noise Jamming (PBNJ)}
PBNJ injects Gaussian noise into a specific sub-band $B_{jam}$. It is generated by filtering complex Gaussian noise $\nu[n]$ given as:
\begin{equation}
    \mathcal{J}_{\text{PBNJ}}[n] = \left( \nu[n] * h_{\text{LPF}}[n] \right) e^{j 2\pi f_c t},
\end{equation}
where $*$ denotes the convolution operation, and $h_{\text{LPF}}$ represents the impulse response of a shaping filter. The jamming bandwidth satisfies $B_{jam} \subset [-F_s/2, F_s/2]$, simulating localized spectral pollution \cite{zhang2011effect}.

\begin{figure}[htbp]
    \centering
    \includegraphics[width=0.95\linewidth]{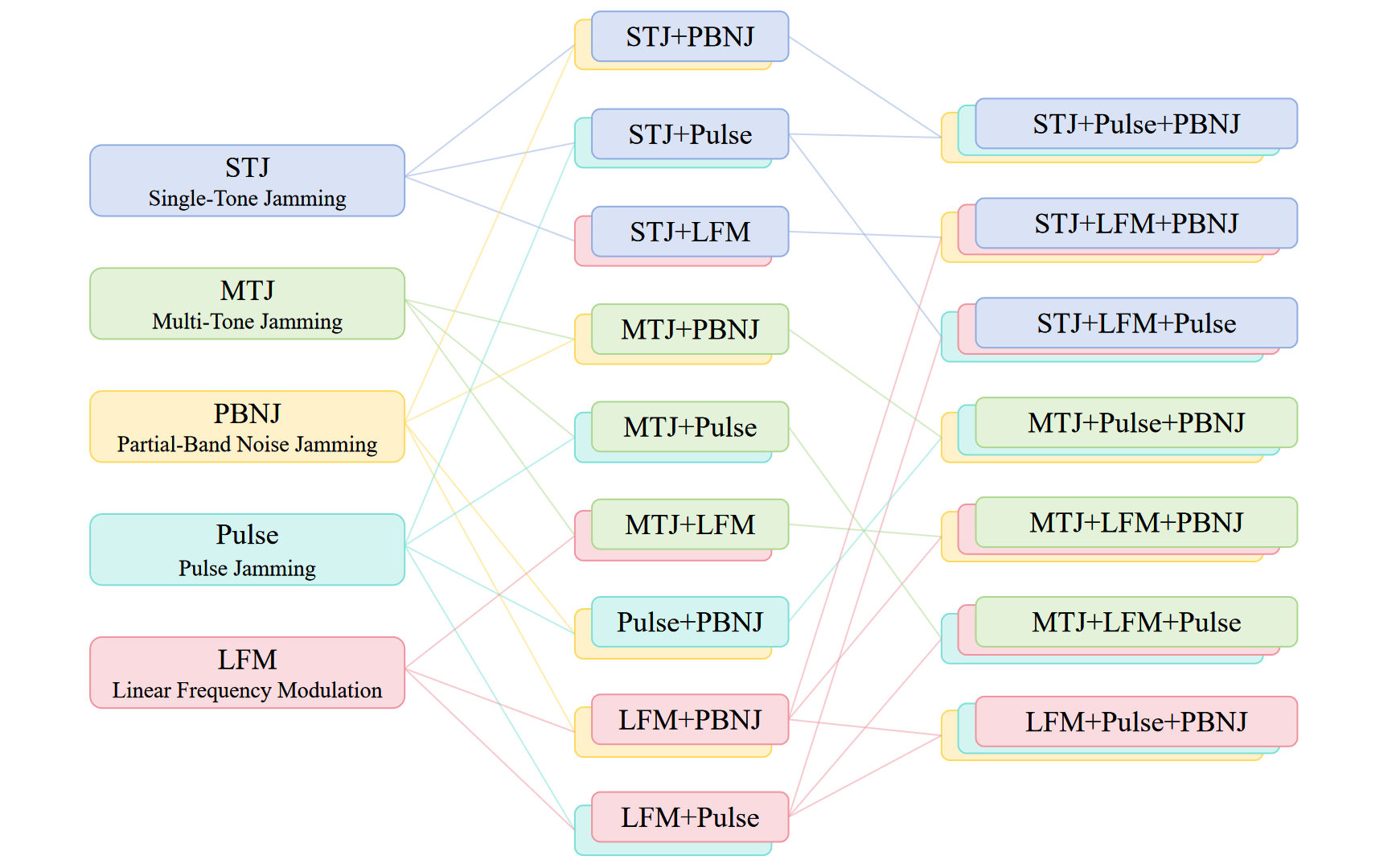}
    \caption{Hierarchical taxonomy of the generated jamming dataset. The model expands from single primitives to dual-component and multi-component compound interference defined by the active set $\mathbb{K}$.}
    \label{fig:jamming_hierarchy}
\end{figure}

\subsection{Compound Jamming Architectures}
\label{subsec:compound_jamming}
Advanced electromagnetic environments employ spatially distributed or multi-mode jammers to create complex interference scenarios \cite{liu2024gnss}. As illustrated in Fig.~\ref{fig:jamming_hierarchy}, we model compound jamming as the sparse superposition of primitives from the set $\Omega$ \cite{xiao2025compound, song2025tfunet}.

Instead of treating compound signals as simple arithmetic additions, we define them formally as a weighted sum of active components. Let $\mathbb{K} \subseteq \Omega$ denote the set of active jamming types for a given scenario (e.g., $\mathbb{K} = \{\text{STJ}, \text{LFM}\}$). The compound interference signal $\mathcal{J}_{\text{compound}}[n]$ is defined as:
\begin{equation}
    \mathcal{J}_{\text{compound}}[n] = \sum_{k \in \mathbb{K}} \sqrt{P_k} \cdot \mathcal{J}_{k}[n],
\end{equation}
where $\mathcal{J}_{k}[n]$ represents the unit-power waveform of the $k$-th primitive, and $P_k$ denotes the power allocated to the $k$-th component.

The total jamming power is given by $P_J = \sum_{k \in \mathbb{K}} P_k$. We define the power distribution among components using the Power Ratio (PR). For a dual-component scenario ($\mathbb{K}=\{A, B\}$), the relationship is controlled by $\text{PR}_{A/B} = 10\log_{10}(P_A/P_B)$.

Based on this formulation, we systematically generate nine specific types of dual-compound interference. This set comprises the superposition of narrowband and sweep/burst signals, specifically STJ+LFM, STJ+Pulse, and STJ+PBNJ; the mixture of multi-tone and non-stationary signals, namely MTJ+LFM, MTJ+Pulse, and MTJ+PBNJ; and the mutual combinations of wideband and dynamic types, including LFM+Pulse, LFM+PBNJ, and Pulse+PBNJ \cite{xiao2025compound}. 

For the highest complexity tier, seven triple-component types are synthesized to represent saturated jamming environments: STJ+LFM+Pulse, STJ+LFM+PBNJ, STJ+Pulse+PBNJ, MTJ+LFM+Pulse, MTJ+LFM+PBNJ, MTJ+Pulse+PBNJ, and the fully dynamic mixture LFM+Pulse+PBNJ. 

Distinct from prior studies \cite{liu2024gnss, mehr2025deep, chen2022fingerprint, li2025dualGCN} which typically focus on single interference sources or a limited subset of composite patterns, our work constructs a significantly more comprehensive library spanning 21 distinct jamming categories, particularly featuring high-complexity triple-component mixtures.

Finally, the total compound interference $\mathcal{J}[n]$ is re-normalized to match the target JNR before being added to the noise floor as defined in Eq. \eqref{eq:rx_signal}.

\begin{figure}[!t]
    \centering
    \begin{subfigure}{0.32\linewidth}
        \includegraphics[width=\linewidth]{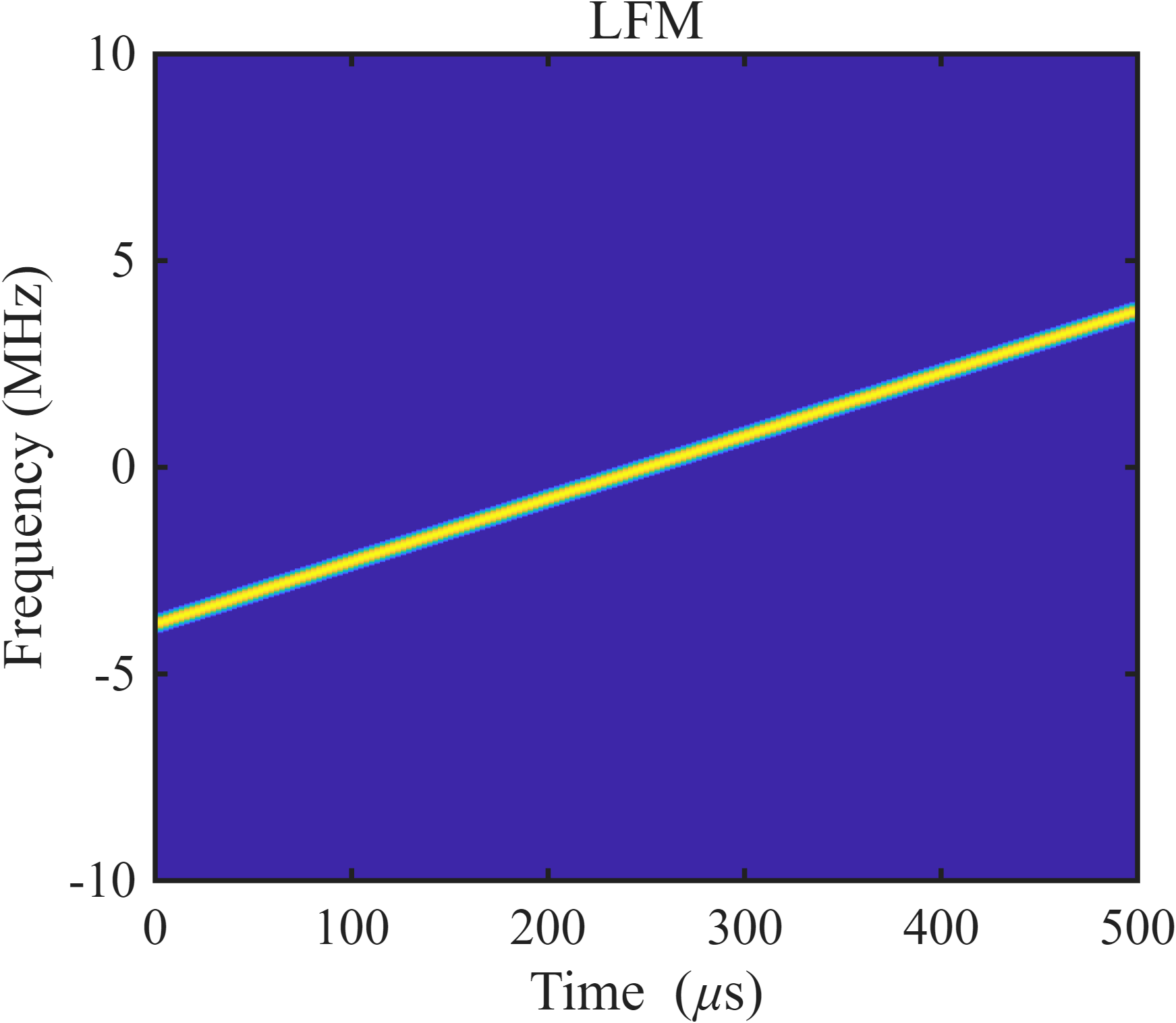}
        \caption{LFM}
        \label{fig:stft_lfm}
    \end{subfigure}
    \hfill
    \begin{subfigure}{0.32\linewidth}
        \includegraphics[width=\linewidth]{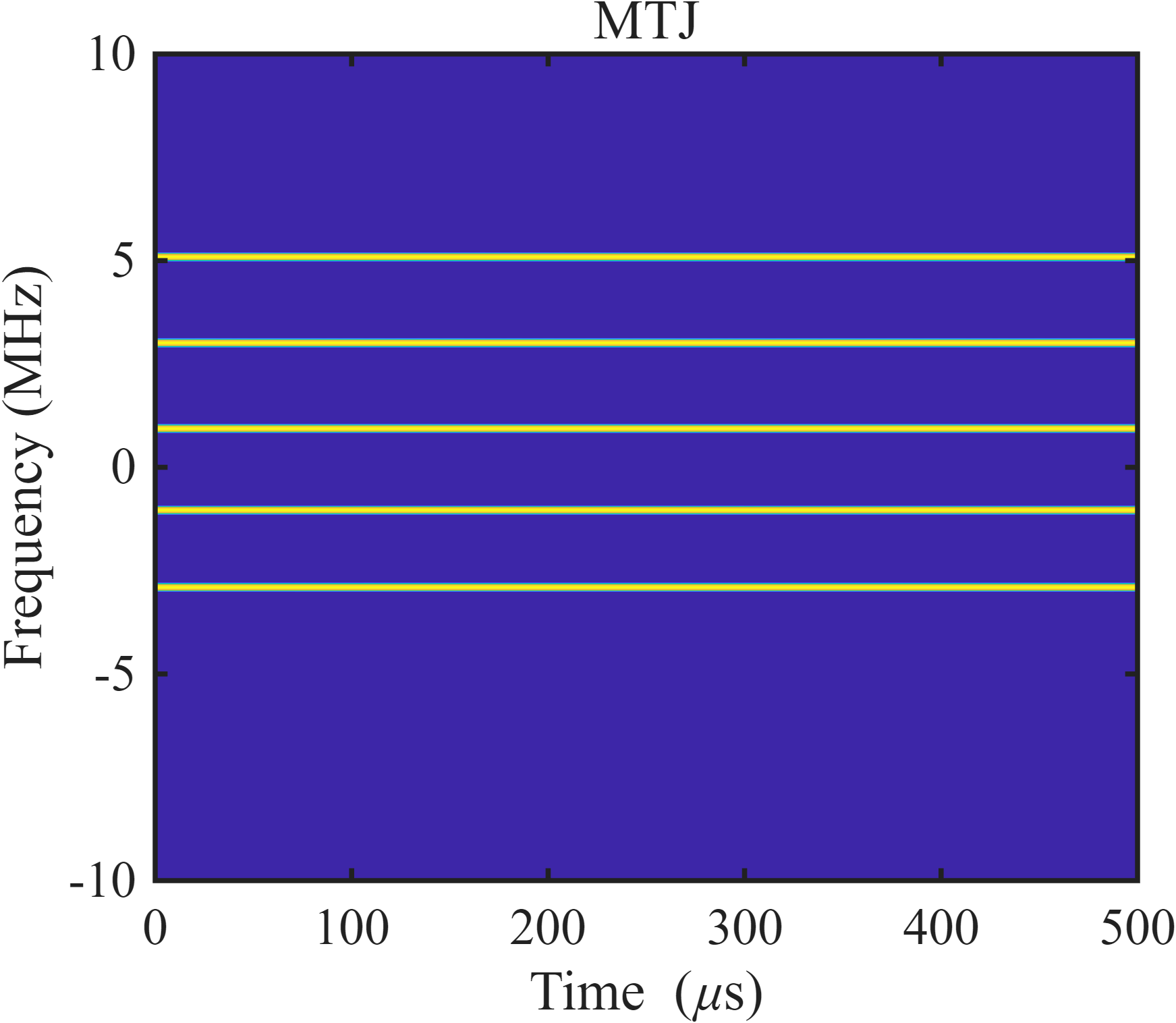}
        \caption{MTJ}
        \label{fig:stft_mtj}
    \end{subfigure}
    \hfill
    \begin{subfigure}{0.32\linewidth}
        \includegraphics[width=\linewidth]{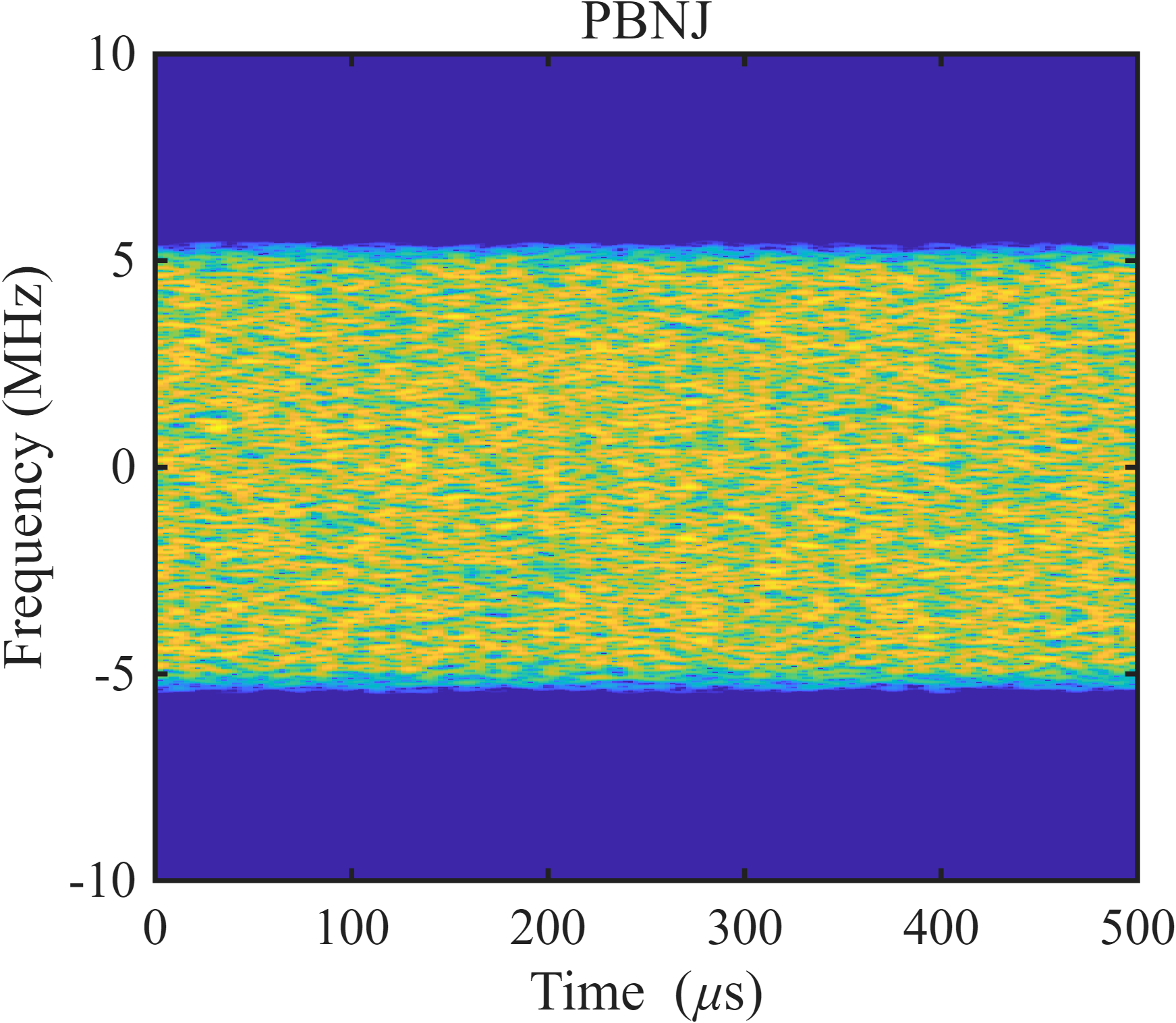}
        \caption{PBNJ}
        \label{fig:stft_pbnj}
    \end{subfigure}
    
    \vspace{0.2cm}
    
    \begin{subfigure}{0.32\linewidth}
        \includegraphics[width=\linewidth]{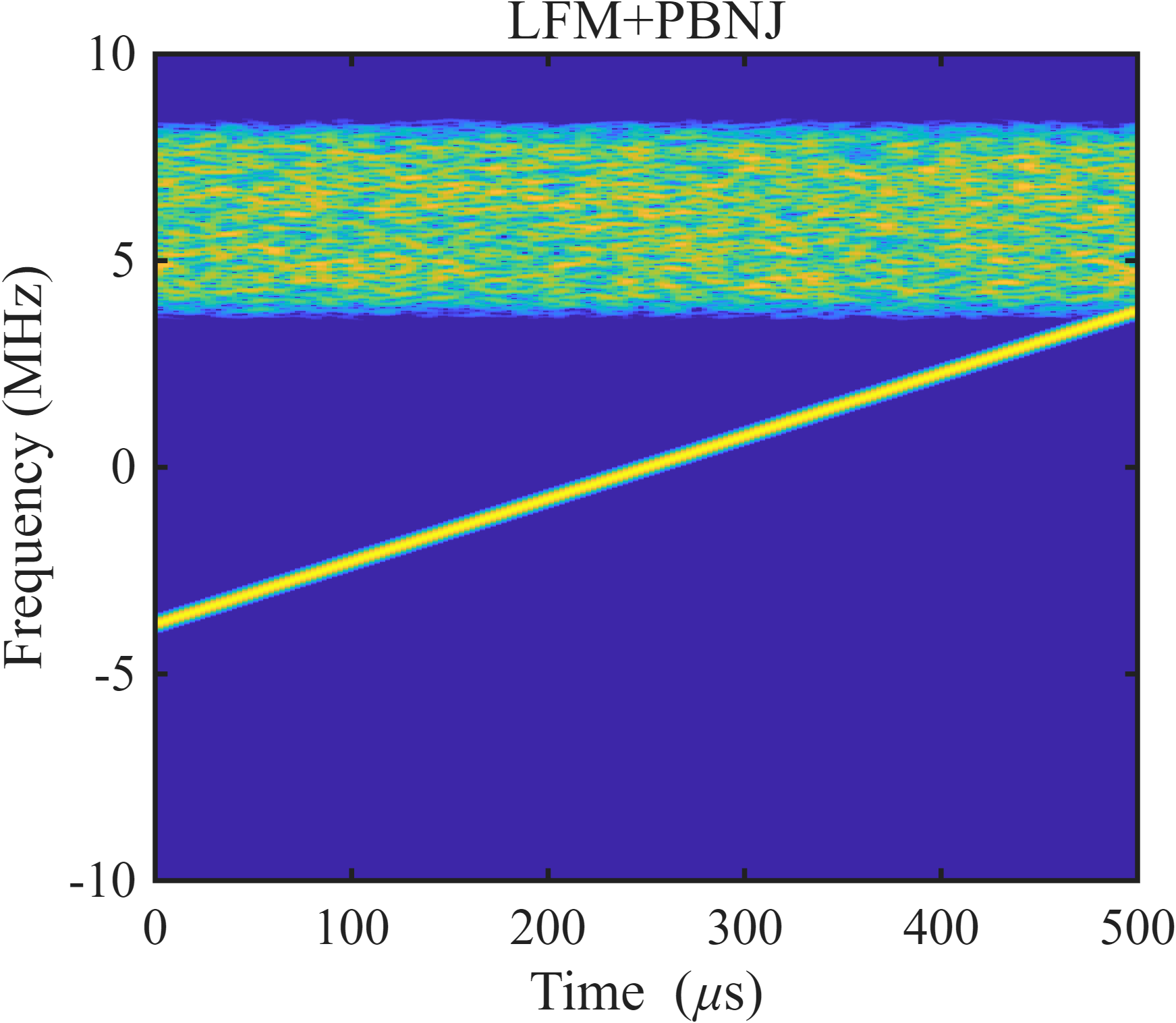}
        \caption{LFM+PBNJ}
        \label{fig:stft_lfm_pbnj}
    \end{subfigure}
    \hfill
    \begin{subfigure}{0.32\linewidth}
        \includegraphics[width=\linewidth]{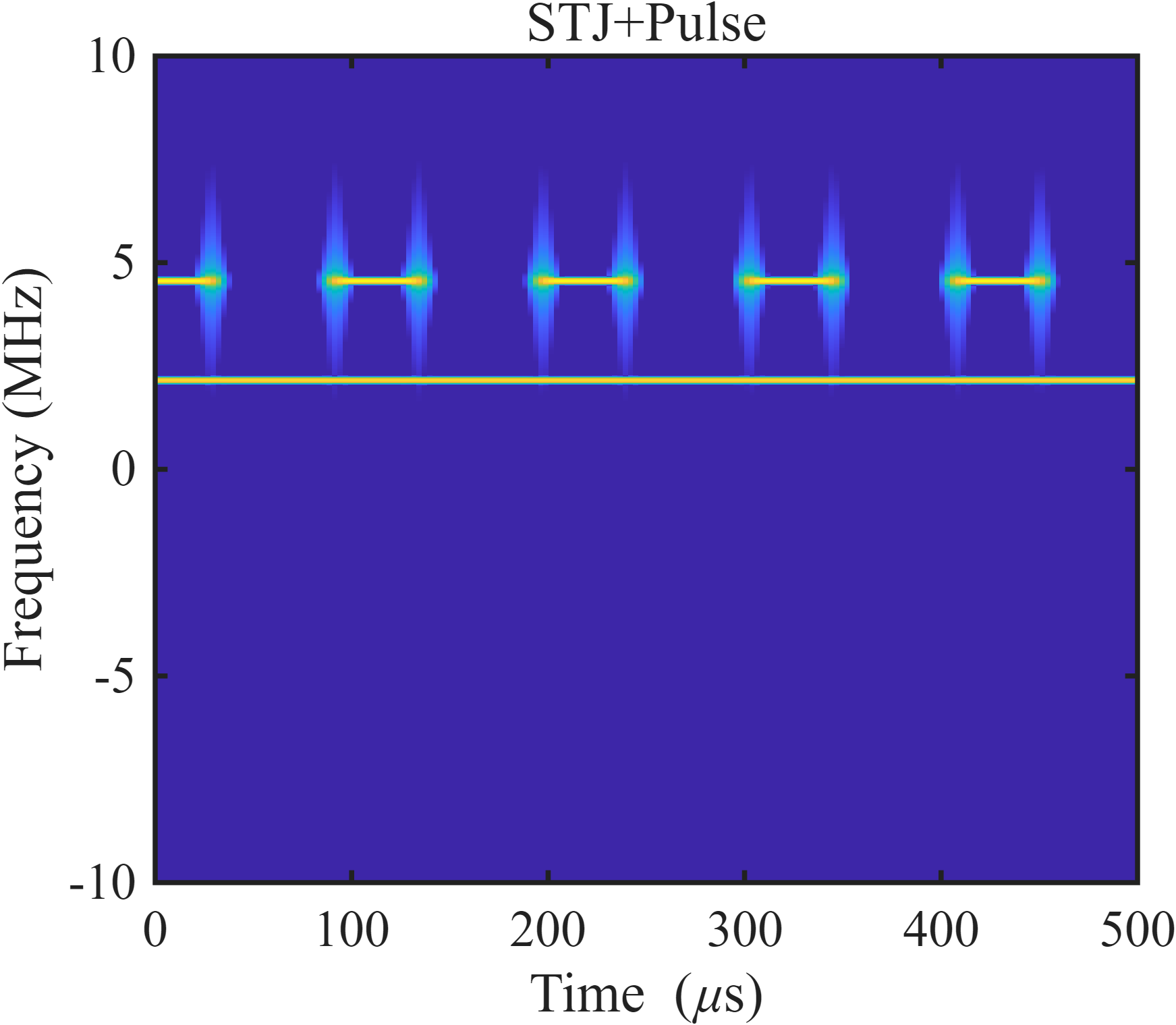}
        \caption{STJ+Pulse}
        \label{fig:stft_stj_pulse}
    \end{subfigure}
    \hfill
    \begin{subfigure}{0.32\linewidth}
        \includegraphics[width=\linewidth]{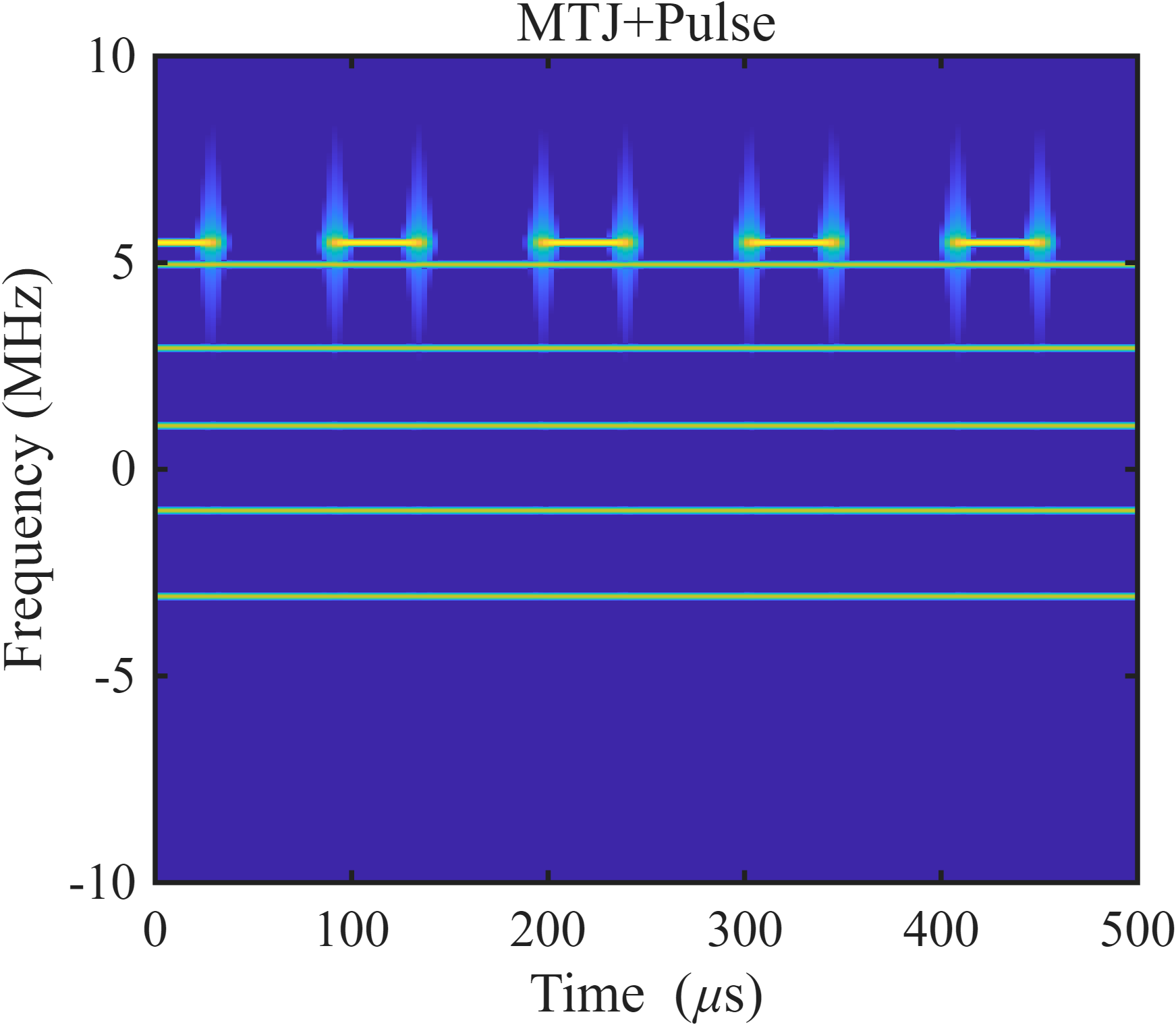}
        \caption{MTJ+Pulse}
        \label{fig:stft_mtj_pulse}
    \end{subfigure}
    
    \vspace{0.2cm}
    
    \begin{subfigure}{0.32\linewidth}
        \includegraphics[width=\linewidth]{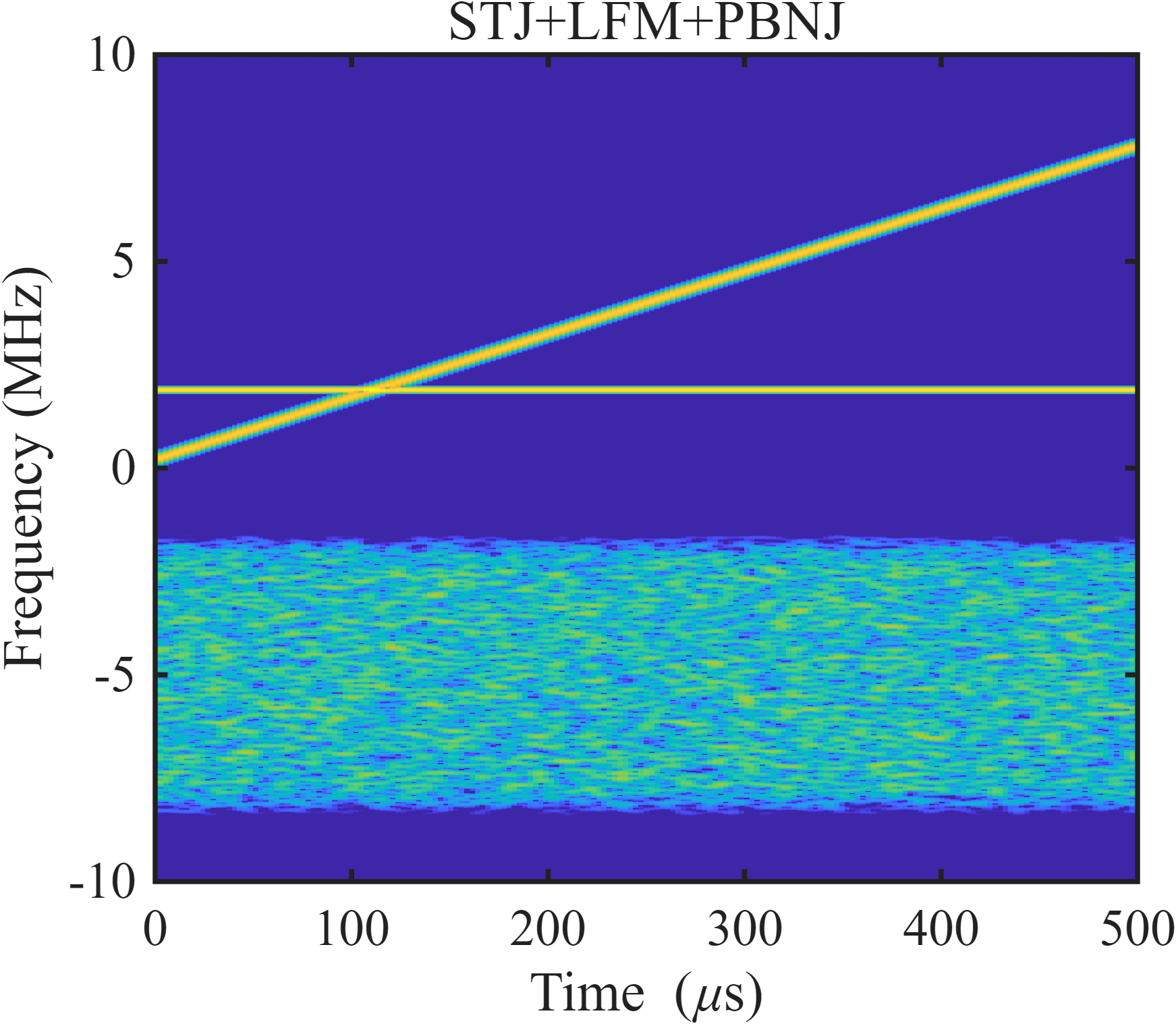}
        \caption{STJ+LFM+PBNJ}
        \label{fig:stft_stj_lfm_pbnj}
    \end{subfigure}
    \hfill
    \begin{subfigure}{0.32\linewidth}
        \includegraphics[width=\linewidth]{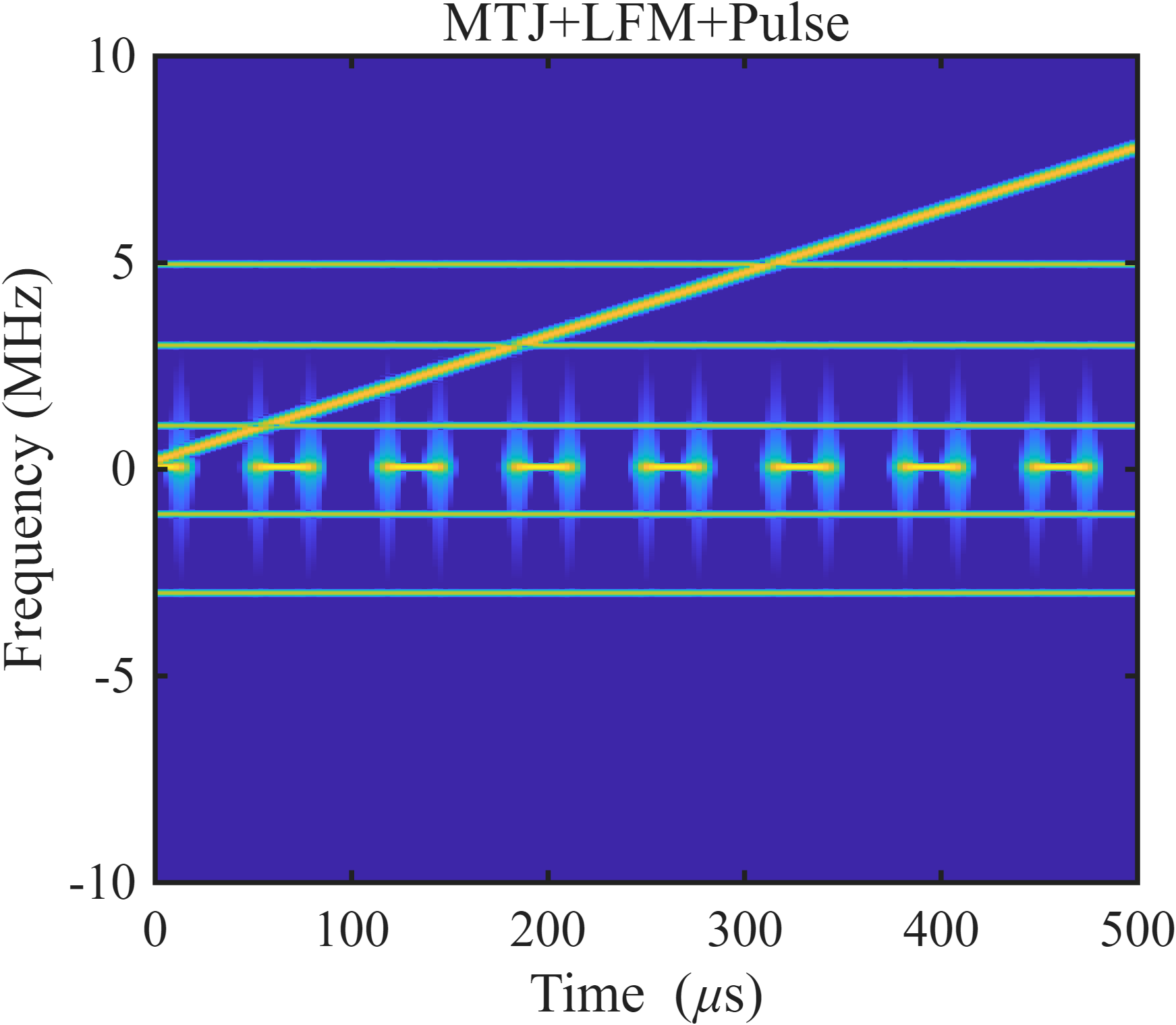}
        \caption{MTJ+LFM+Pulse}
        \label{fig:stft_mtj_lfm_pulse}
    \end{subfigure}
    \hfill
    \begin{subfigure}{0.32\linewidth}
        \includegraphics[width=\linewidth]{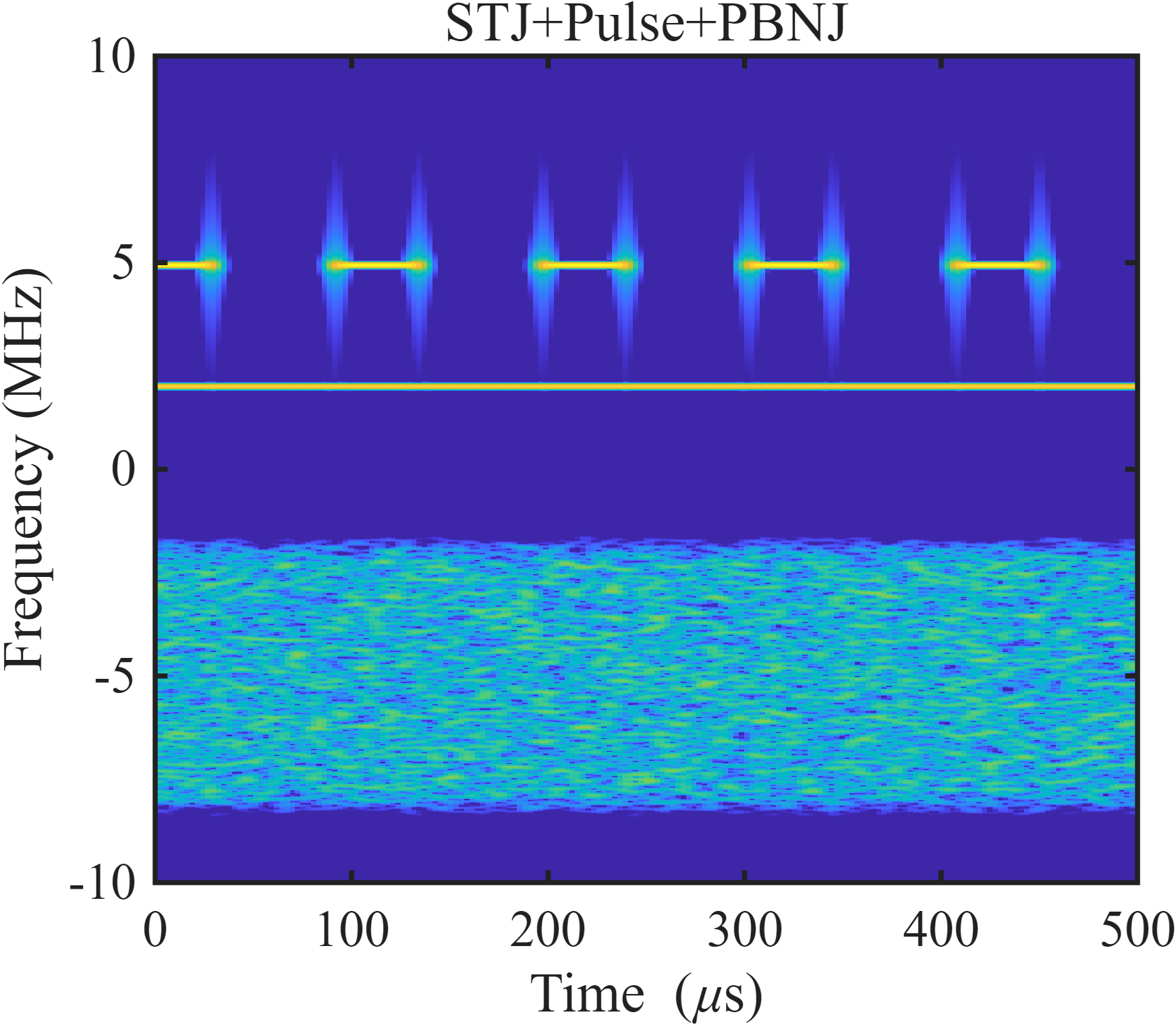}
        \caption{STJ+Pulse+PBNJ}
        \label{fig:stft_stj_pulse_pbnj}
    \end{subfigure}
    
    \caption{STFT Spectrograms of different jamming scenarios. Top row: Single jamming primitives. Middle row: Dual compound jamming. Bottom row: Triple compound jamming. The TF representation effectively reveals the complex entanglement of features, such as the vertical bursts of Pulse jamming intersecting with the continuous sweep of LFM, which challenges static classifiers.}
    \label{fig:stft_matrix}
\end{figure}

\section{Signal Analysis}
\label{sec:signal_analysis}

To effectively discriminate between diverse jamming primitives and their compound superpositions, it is essential to extract features capable of capturing both spectral occupancy and temporal dynamics \cite{xiao2025compound, wang2018detection}. Accordingly, the raw time-domain signal $x[n]$ is transformed into two complementary feature domains: the TF domain via the STFT and the frequency domain via the PSD.

\begin{figure}[!t]
    \centering
    \begin{subfigure}{0.32\linewidth}
        \includegraphics[width=\linewidth]{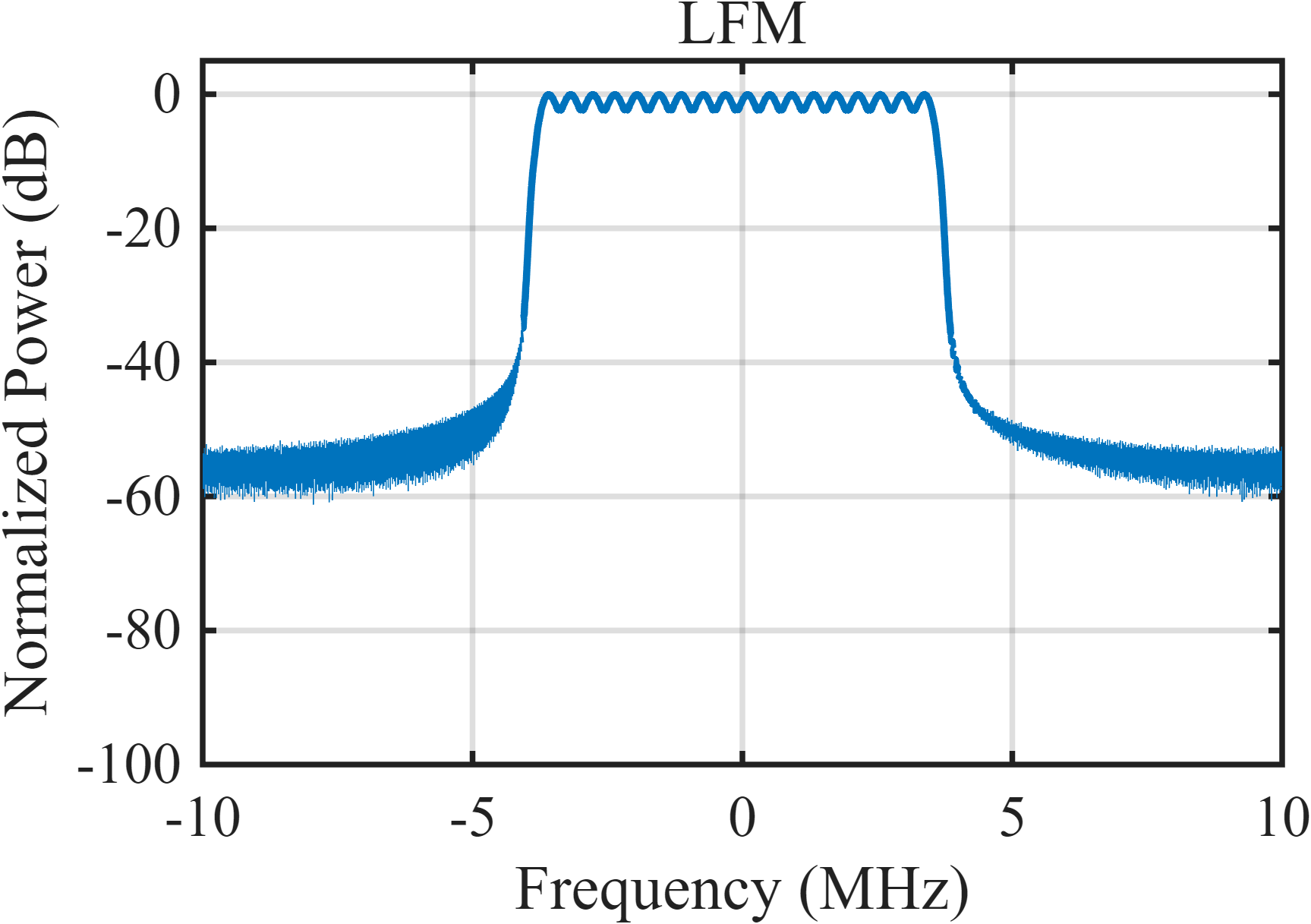} 
        \caption{LFM}
    \end{subfigure}
    \hfill
    \begin{subfigure}{0.32\linewidth}
        \includegraphics[width=\linewidth]{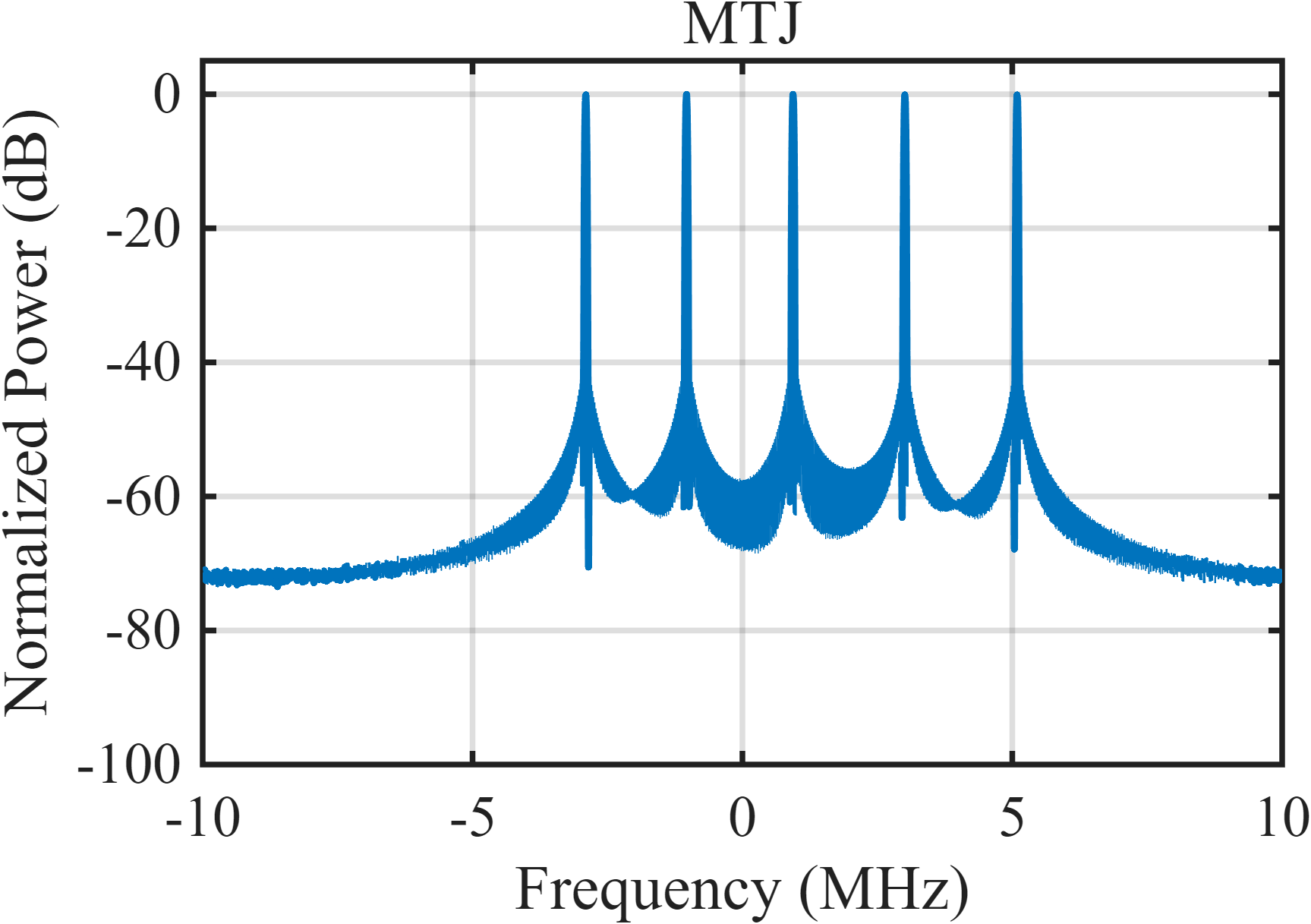}
        \caption{MTJ}
    \end{subfigure}
    \hfill
    \begin{subfigure}{0.32\linewidth}
        \includegraphics[width=\linewidth]{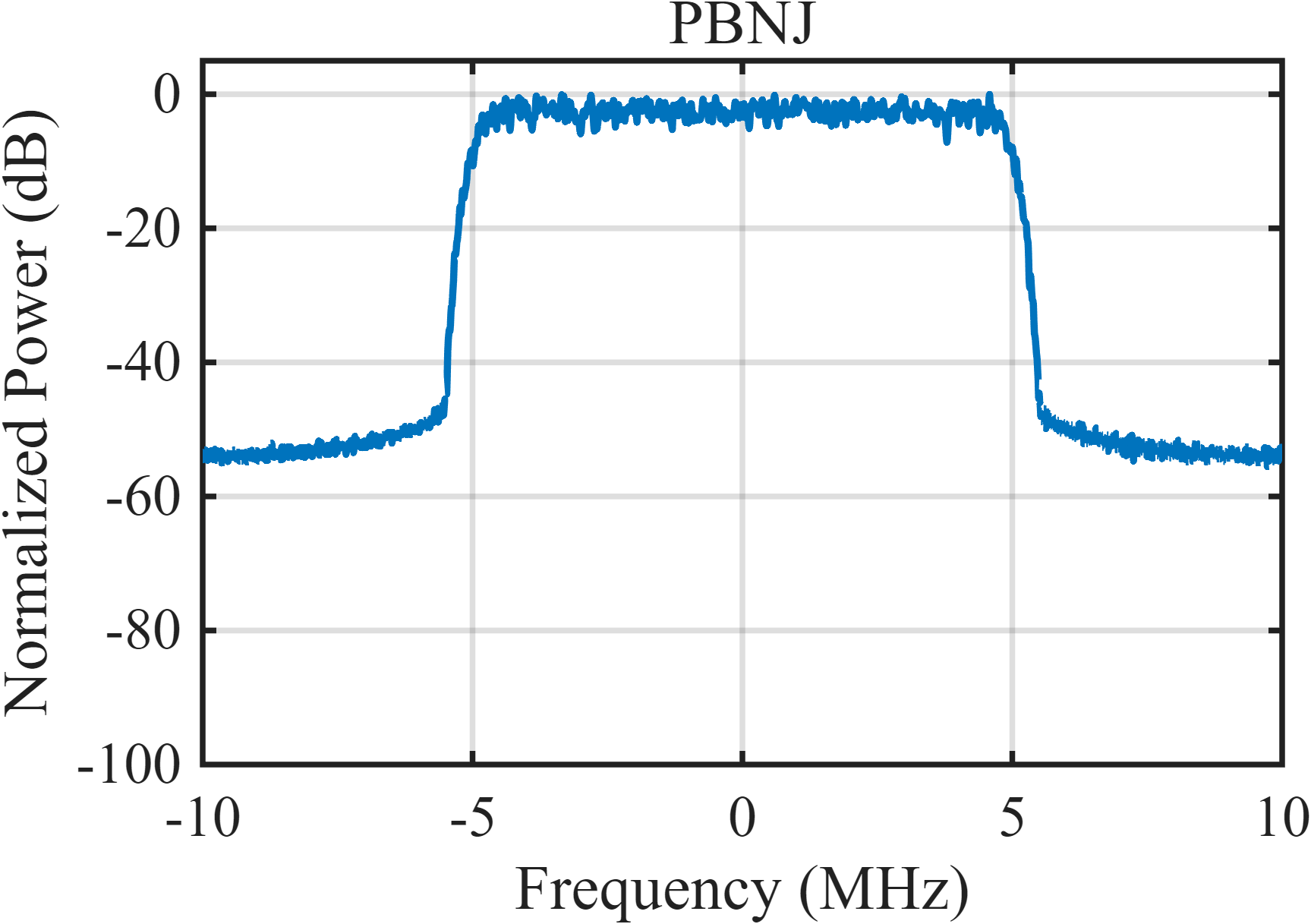} 
        \caption{PBNJ}
    \end{subfigure}
    
    \vspace{0.75cm} 
    
    \begin{subfigure}{0.32\linewidth}
        \includegraphics[width=\linewidth]{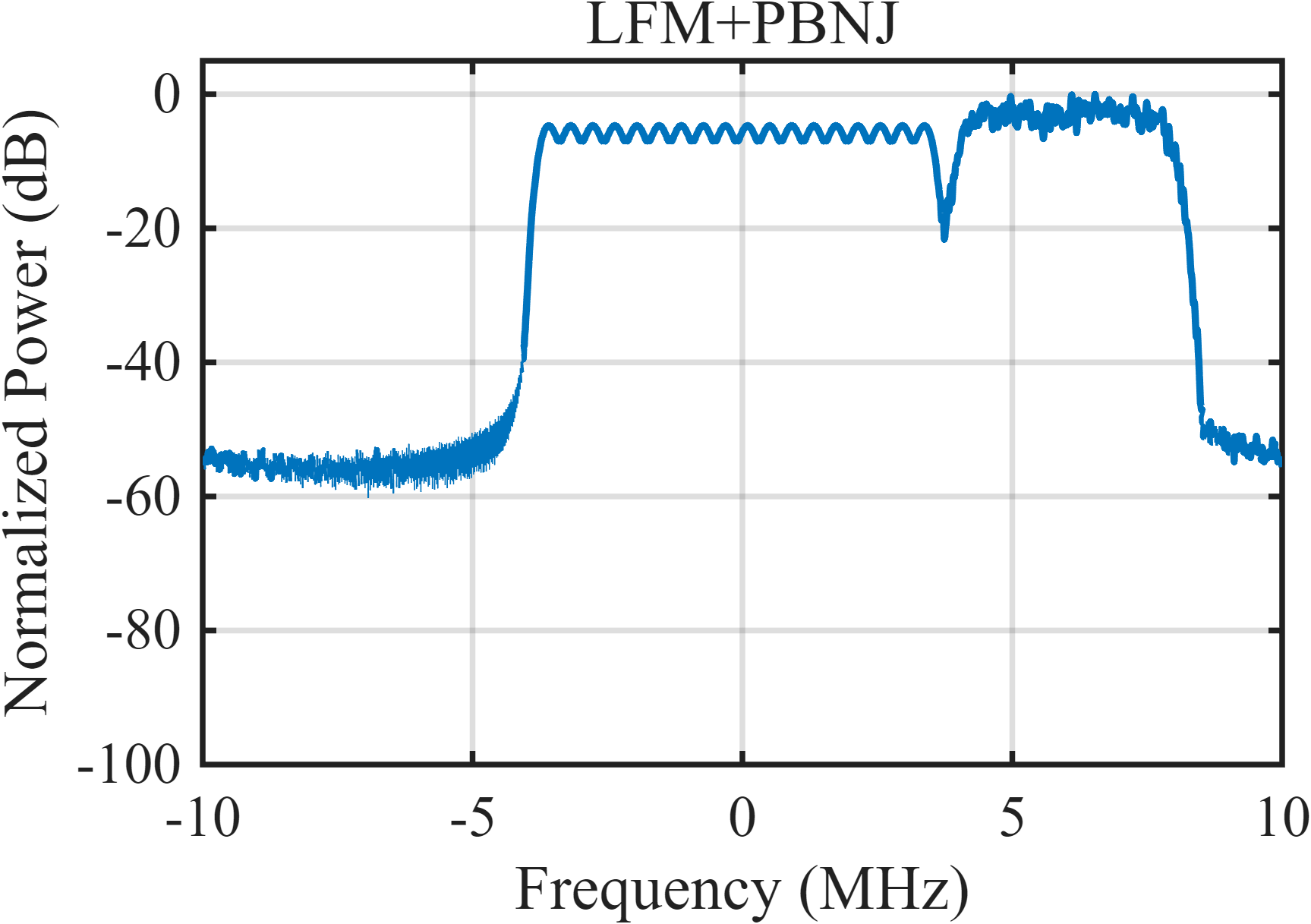} 
        \caption{LFM+PBNJ}
    \end{subfigure}
    \hfill
    \begin{subfigure}{0.32\linewidth}
        \includegraphics[width=\linewidth]{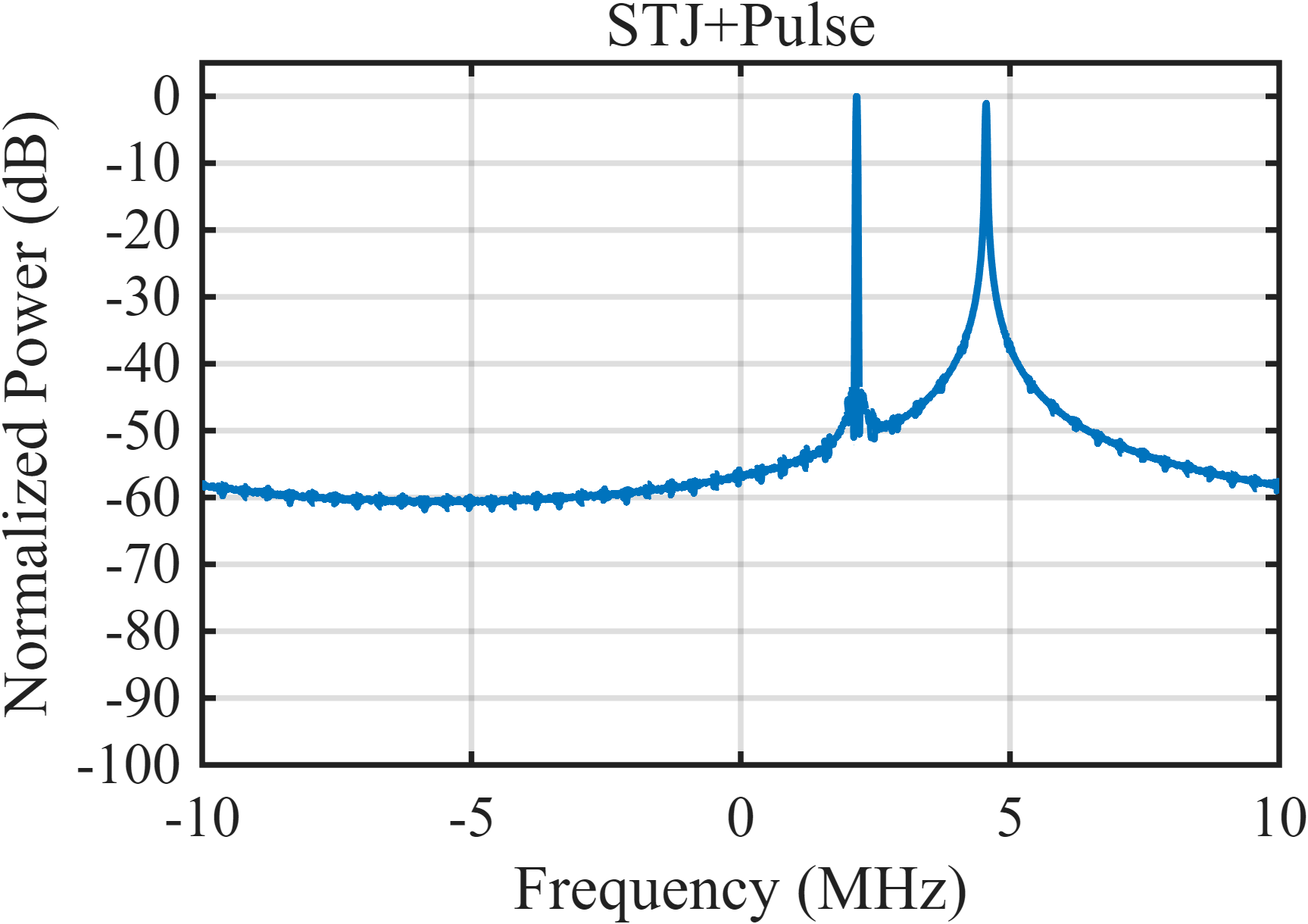}
        \caption{STJ+Pulse}
    \end{subfigure}
    \hfill
    \begin{subfigure}{0.32\linewidth}
        \includegraphics[width=\linewidth]{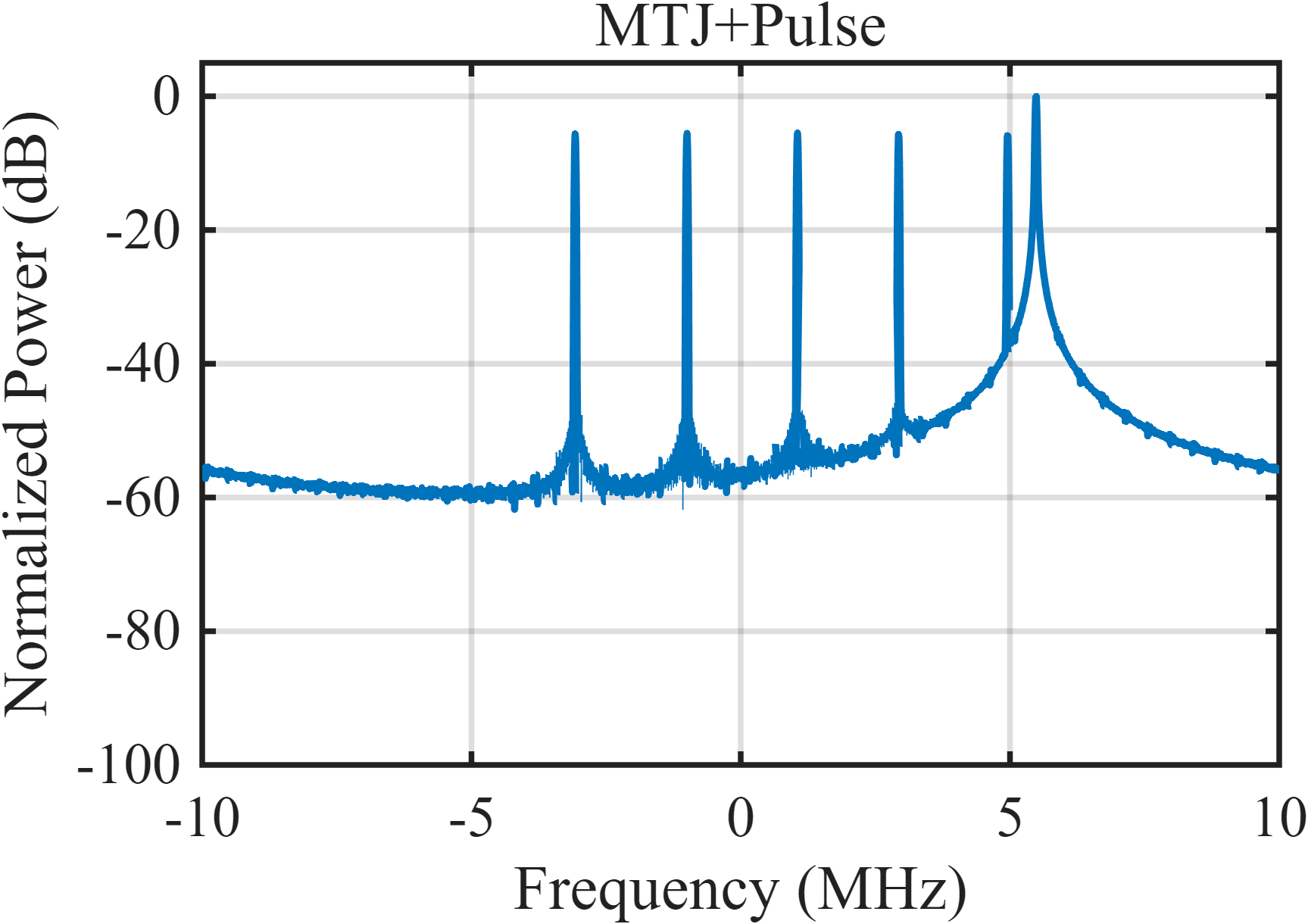}
        \caption{MTJ+Pulse}
    \end{subfigure}
    
    \vspace{0.75cm} 
    
    \begin{subfigure}{0.32\linewidth}
        \includegraphics[width=\linewidth]{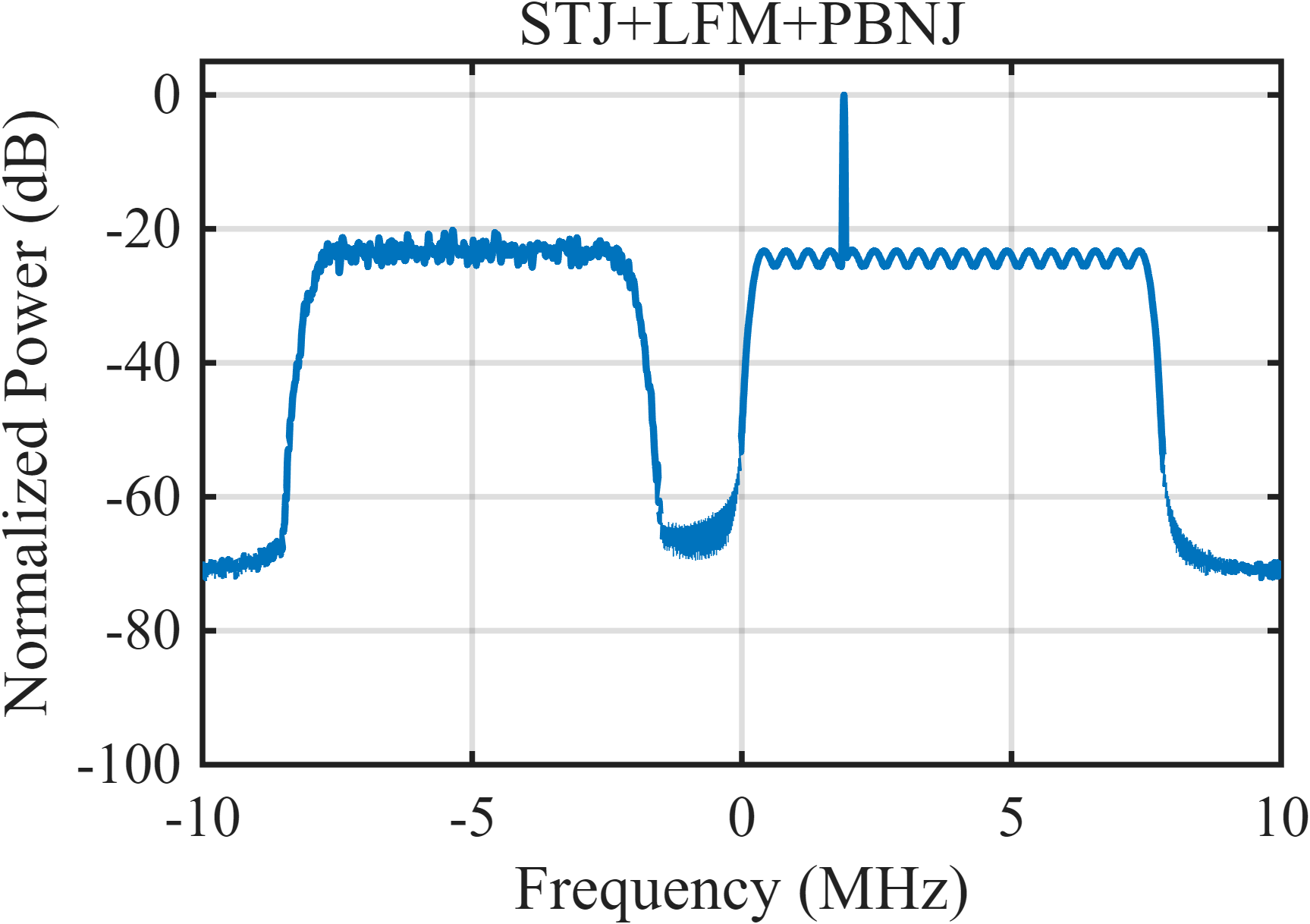} 
        \caption{STJ+LFM+PBNJ}
    \end{subfigure}
    \hfill
    \begin{subfigure}{0.32\linewidth}
        \includegraphics[width=\linewidth]{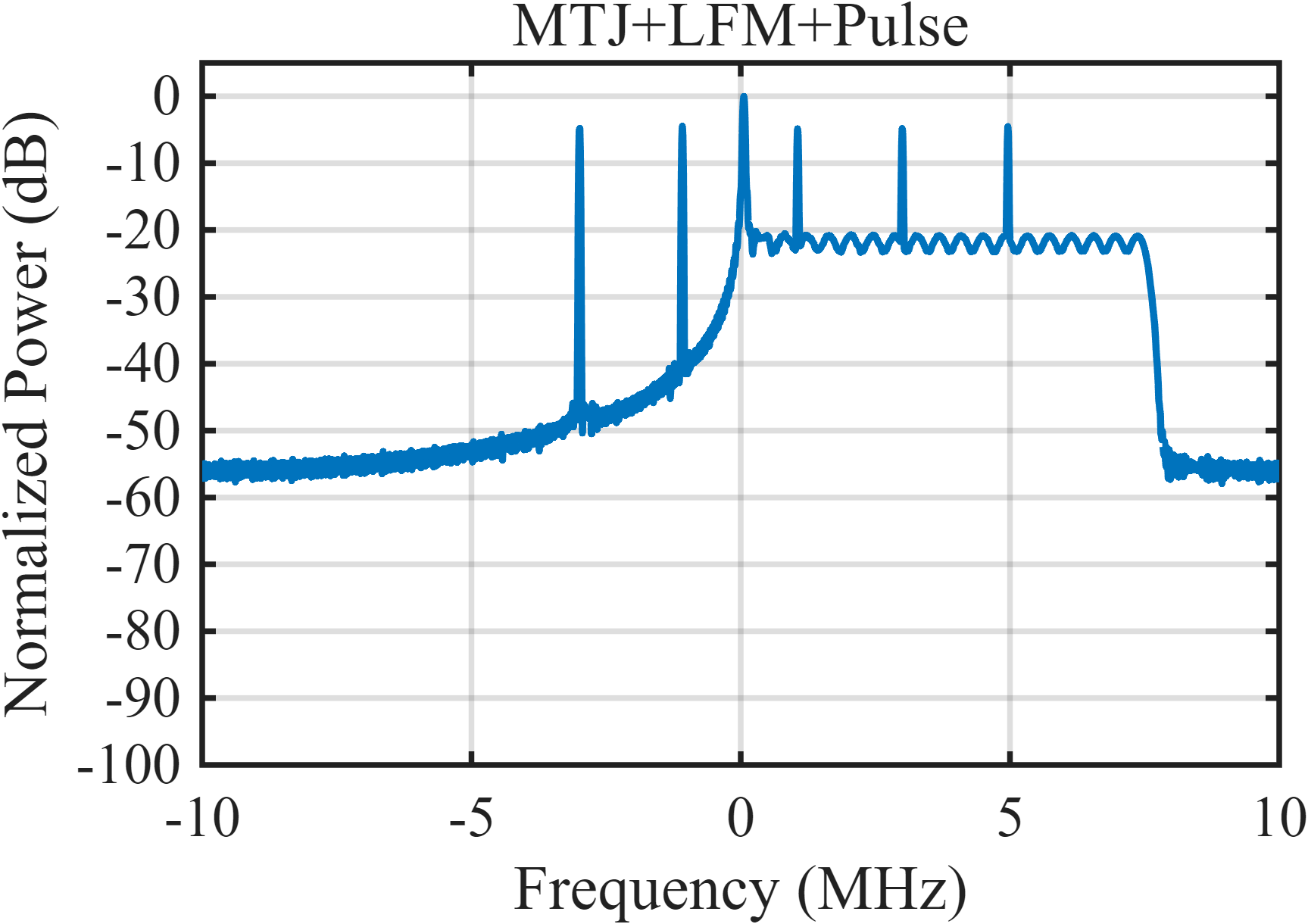}
        \caption{MTJ+LFM+Pulse}
    \end{subfigure}
    \hfill
    \begin{subfigure}{0.32\linewidth}
        \includegraphics[width=\linewidth]{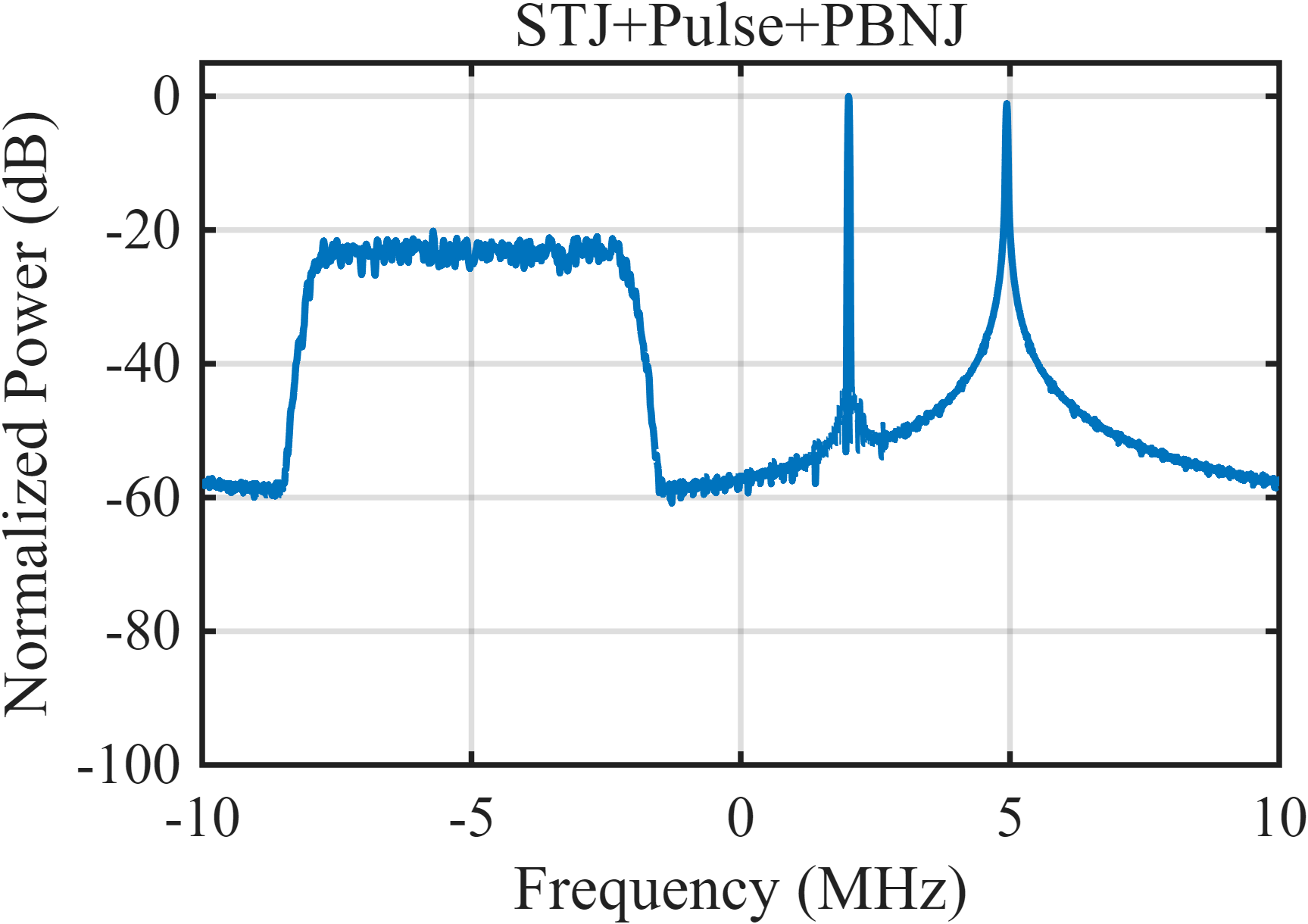}
        \caption{STJ+Pulse+PBNJ}
    \end{subfigure}
    
    \caption{Power Spectral Density (PSD) estimates of the corresponding jamming scenarios. The MTM-based PSD provides a clear view of the energy distribution across frequencies, aiding in the differentiation of spectrally overlapping signals.}
    \label{fig:psd_matrix}
\end{figure}

\subsection{Time-Frequency Representation via STFT}
The STFT characterizes the evolution of the signal's frequency content over time, which is critical for identifying non-stationary interference patterns, including the chirp rate of LFM or the periodicity of Pulse jamming \cite{cai2019waterfall, sun2024frft}.

The discrete-time signal $x[n]$ of length $N$ is analyzed using a sliding window approach. A Hann window $w_{\text{Hann}}[n]$ is employed to mitigate spectral leakage. The window function, the discrete STFT $X[m, k]$, and the standard power spectrogram $S[m, k]$ are defined as follows:
\begin{align}
    w_{\text{Hann}}[n] &= \sin^2\left(\frac{\pi n}{N_{\text{win}}-1}\right), \quad 0 \le n < N_{\text{win}}, \\
    X[m, k] &= \sum_{n=0}^{N-1} x[n] w_{\text{Hann}}[n - mR] e^{-j \frac{2\pi}{N_{\text{fft}}} k n}, \\
    S[m, k] &= \left| X[m, k] \right|^2, \label{eq:power_spectrogram}
\end{align}
where $m$ denotes the time frame index, $k$ is the frequency bin index ($0 \le k < N_{\text{fft}}$), $N_{\text{win}}$ is the window length, $N_{\text{fft}}$ is the FFT size, and $R$ represents the hop size.

Although Eq. \eqref{eq:power_spectrogram} represents the signal power, effectively identifying secondary components within compound jamming scenarios necessitates enhanced contrast for low-amplitude features. Consequently, instead of directly employing the power spectrum, we perform non-linear dynamic range compression on the magnitude spectrum $|X[m,k]|$. The final log-magnitude TF feature, $I_{\text{TF}}[m, k]$, is given by
\begin{equation}
    I_{\text{TF}}[m, k] = 20 \log_{10} \left( \frac{|X[m, k]|^{\gamma}}{\max(|X[m, k]|^{\gamma}) + \epsilon} \right),
    \label{eq:stft_log_mag}
\end{equation}
where $\gamma$ is the gamma correction factor for contrast enhancement, and $\epsilon$ is a regularization constant ensuring numerical stability. The factor of 20 corresponds to the logarithmic scaling of the signal magnitude. The resulting matrix is resized to dimensions $H \times W$ and pseudo-colored to serve as the input for the deep learning classifier \cite{mehr2025gru, song2025tfunet}. Representative STFT spectrograms for single, dual, and triple-component jamming scenarios are visualized in Fig. \ref{fig:stft_matrix}.

\subsection{Spectral Analysis via Multitaper Method}
In contrast to the STFT, which captures temporal dynamics, the PSD provides a high-fidelity estimation of the energy distribution across the frequency spectrum \cite{wang2017statistical}. To obtain a spectral estimate with reduced variance while maintaining frequency resolution—a limitation often encountered in classical periodogram or Welch's methods—we employ the Multitaper Method (MTM).

The MTM reduces the variance of the spectral estimate by utilizing a set of $N_{tapers}$ orthogonal data tapers, known as Discrete Prolate Spheroidal Sequences (DPSS) or Slepian sequences. Notably, Slepian sequences are optimized to maximize energy concentration within the central lobe, thereby significantly mitigating spectral leakage \cite{xiao2025compound}. This suppression is indispensable in compound jamming scenarios, where the sidelobes of a dominant high-power component might otherwise obscure a weaker superimposed signal in conventional spectral estimates.

Let $x[n]$ denote the discrete-time signal of length $N$. We define a set of $N_{tapers}$ orthonormal tapers $v_p[n]$ ($p=0, \dots, N_{tapers}-1$). The eigencoefficients $Y_p(f)$ for the $p$-th taper and the final Multitaper PSD estimate $\hat{P}_{\text{MTM}}(f)$ are formulated as follows:
\begin{align}
    \sum_{n=0}^{N-1} v_p[n] v_j[n] &= \delta_{p,j}, \\
    Y_p(f) &= \sum_{n=0}^{N-1} x[n] v_p[n] e^{-j 2\pi f n}, \\
    \hat{P}_{\text{MTM}}(f) &= \frac{1}{N_{tapers}} \sum_{p=0}^{N_{tapers}-1} \left| Y_p(f) \right|^2,
\end{align}
where $\delta_{p,j}$ represents the Kronecker delta. Typically, $N_{tapers}$ is chosen as ($2N\cdot W_{bw} - 1$) to balance bias and variance, where $N\cdot W_{bw}$ is the time-bandwidth product, $W_{bw}$ is the normalized half-bandwidth resolution.

Similar to the TF representation, the estimated PSD is converted to a logarithmic scale to facilitate the detection of weak interference components. The final spectral feature vector $I_{\text{PSD}}(f)$ is given by
\begin{equation}
    I_{\text{PSD}}(f) = 10 \log_{10} \left( \hat{P}_{\text{MTM}}(f) \right).
\end{equation}
This spectral curve provides a robust frequency-domain signature complementary to the STFT spectrogram. Unlike the STFT, which preserves local temporal dynamics, the PSD offers a compact, shift-invariant representation of global spectral complexity. Therefore, it serves as a robust indicator for the proposed physics-guided gating mechanism, enabling the system to optimally allocate computational resources based on the spectral characteristics of the jamming environment \cite{xiao2025compound}. The corresponding PSD estimates generated via MTM are illustrated in Fig. \ref{fig:psd_matrix}.

\section{Methodology}
\label{sec:methodology}

\begin{figure*}[htbp]
    \centering
    \includegraphics[width=0.95\textwidth]{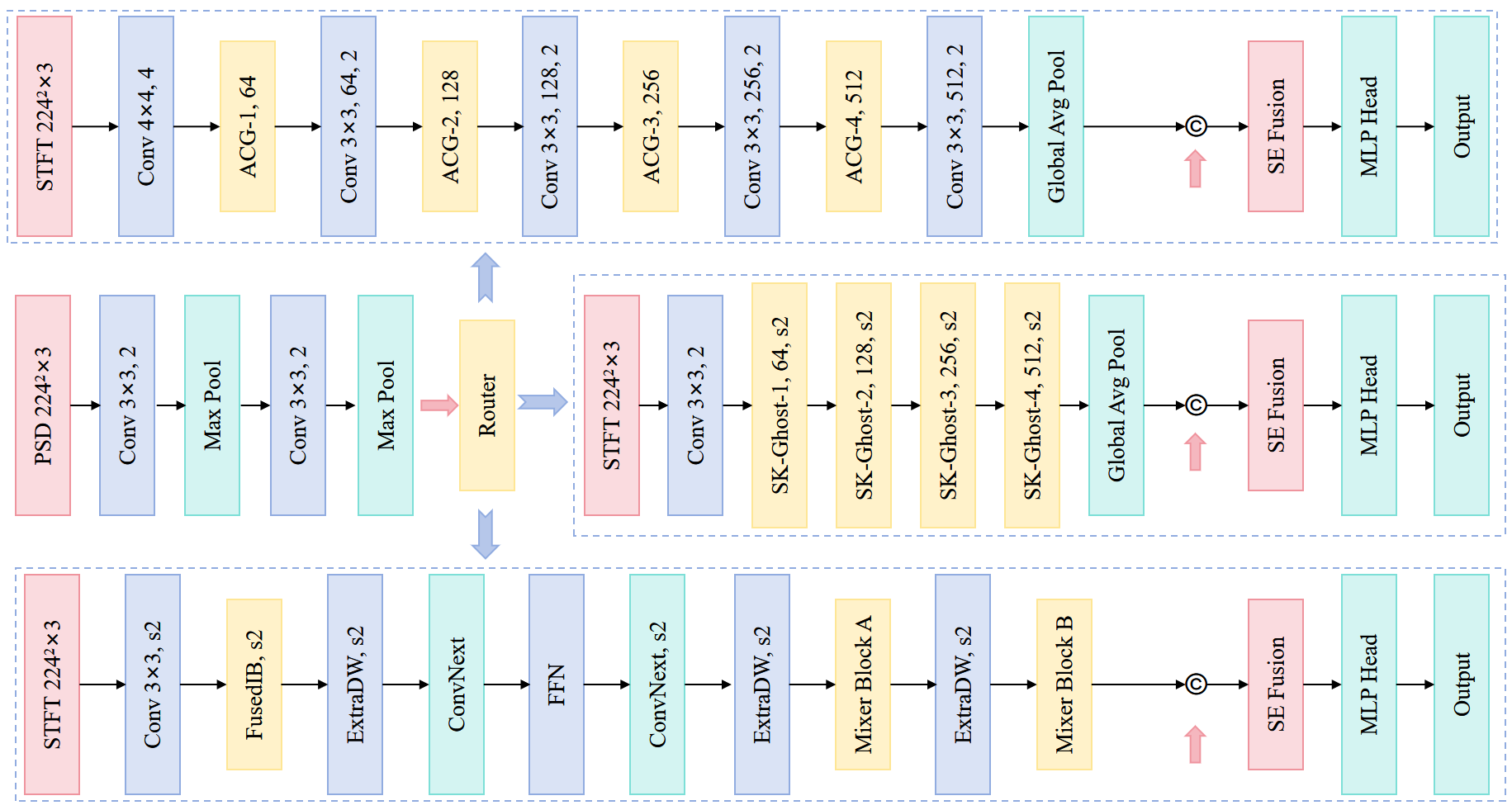}
    \caption{Overview of the Proposed Cognitive Dual-Stream MoE Framework. The architecture integrates three specialized expert networks: (1) A TransNeXt-based expert with Coordinate-GLU for directional feature modeling in saturated environments, (2) An SK-GhostNet expert for multi-scale adaptive reception under varying bandwidths, and (3) A MobileNetV4 expert utilizing Universal Inverted Bottlenecks (UIB) for hardware-aware universal feature extraction. A spectrum-aware router mechanism dynamically aggregates expert outputs for robust classification.}
    \label{fig:moe_framework}
\end{figure*}

In this study, we propose a robust interference classification framework based on an MoE architecture \cite{an2025moe, xu2023slmoe}, as illustrated in Fig.~\ref{fig:moe_framework}. To effectively handle the complex TF characteristics of GNSS interference signals \cite{liu2024gnss, mehr2025deep}, the proposed framework integrates three specialized expert networks, each designed to extract complementary features corresponding to different signal complexities. In this section, we first detail the problem formulation, followed by the specific designs of the expert networks and the physics-guided gating mechanism.

\subsection{Problem Formulation}

Given a coherent observation window, the GNSS receiver captures a discrete-time complex baseband sequence $r[n]$. To construct a robust feature space for interference recognition, this sequence is mapped into dual-domain representations: the TF spectrogram $\mathbf{X}_{\text{TF}} \in \mathbb{R}^{H \times W}$ and the Power Spectral Density image $\mathbf{X}_{\text{PSD}} \in \mathbb{R}^{H \times W}$ \cite{xiao2025compound}. Consequently, the input space is defined as a dataset $\mathcal{D} = \{(\mathbf{X}^{(i)}, y^{(i)})\}_{i=1}^{N_{data}}$, where $\mathbf{X}^{(i)} = \{\mathbf{X}_{\text{TF}}^{(i)}, \mathbf{X}_{\text{PSD}}^{(i)}\}$ denotes the feature pair of the $i$-th sample, and $y^{(i)} \in \{1, \dots, C\}$ represents the corresponding ground-truth label among $C$ jamming categories.

The primary objective is to approximate the optimal mapping function $\mathcal{M}: \mathcal{X} \rightarrow \mathcal{Y}$ by learning a parameterized deep neural network model $\mathcal{F}(\mathbf{X}; \Theta)$. Unlike static architectures where parameters are shared globally across all inputs, the proposed PhyG-MoE framework formulates the predictive output $\hat{\mathbf{y}}$ as a dynamic conditional computation \cite{lepikhin2020gshard, an2025moe, xu2023slmoe}.  Mathematically, the output is derived from a convex combination of $N_{E}$ specialized expert networks given as follows:
\begin{equation}
\mathcal{F}(\mathbf{X}; \Theta) = \sum_{e=1}^{N_{E}} \mathcal{G}(\mathbf{X}_{\text{PSD}}; \theta_g)_e \cdot \mathcal{E}_e(\mathbf{X}_{\text{TF}}; \theta_e),
\label{eq:moe_output}
\end{equation}
where $\Theta = \{\theta_g, \theta_1, \dots, \theta_{N_{E}}\}$ encompasses the global trainable parameters. Here, $\mathcal{E}_e(\cdot)$ denotes the output vector of the $e$-th expert network processing the TF spectrogram, and $\mathcal{G}(\cdot)_e$ represents the gating probability assigned to the $e$-th expert, conditioned on the spectral features $\mathbf{X}_{\text{PSD}}$.

It is important to note the distinction between training and inference phases: while (Eq. \eqref{eq:moe_output}) utilizes a soft-gating mechanism during training to ensure differentiability, a Top-1 hard-gating strategy is employed during inference. This ensures that only the single most relevant expert is activated, thereby minimizing computational redundancy.

To estimate the optimal parameters $\Theta^*$, we formulate a composite objective function. Standard MoE training is prone to expert collapse, where the gating network converges to a ``Winner-takes-all" state by consistently activating a single expert for all inputs. To mitigate this and ensure diverse expert utilization, we introduce an auxiliary Load Balancing Loss $\mathcal{L}_{aux}$, adopting a strategy similar to those successfully employed in large-scale sparse modeling \cite{lepikhin2020gshard, xu2023slmoe}. Let $\mathcal{B}$ denote a training batch of size $B$. For each expert $e$, we define the fraction of samples assigned to it, $f_e$, the average gating probability, $\bar{g}_e$, with $\mathbb{I}(\cdot)$ denoting the indicator function, and the resulting auxiliary loss as follows:
\begin{align}
    f_e &= \frac{1}{B} \sum_{\mathbf{X} \in \mathcal{B}} \mathbb{I}\left(e = \mathop{\arg\max}_j \mathcal{G}(\mathbf{X})_j\right), \\
    \bar{g}_e &= \frac{1}{B} \sum_{\mathbf{X} \in \mathcal{B}} \mathcal{G}(\mathbf{X})_e, \\
    \mathcal{L}_{aux}(\Theta) &= N_{E} \sum_{e=1}^{N_{E}} f_e \cdot \bar{g}_e.
\end{align}

Consequently, the total empirical risk $\mathcal{L}_{total}$ is a weighted sum of the Categorical Cross-Entropy (CCE) loss and the auxiliary loss. The total loss and the optimization problem are formulated as follows:
\begin{align}
    \mathcal{L}_{total}(\Theta) &= \mathcal{L}_{\text{CE}}(\Theta) + \lambda \mathcal{L}_{aux}(\Theta), \label{eq:total_loss} \\
    \Theta^* &= \mathop{\arg\min}_{\Theta} \frac{1}{N_{train}} \sum_{i=1}^{N_{train}} \mathcal{L}_{total}(y^{(i)}, \mathcal{F}(\mathbf{X}^{(i)}; \Theta)),
\end{align}
where $\mathcal{L}_{\text{CE}}$ is the standard cross-entropy loss and $\lambda$ is a hyperparameter balancing classification accuracy and expert load distribution. The optimization problem is finally solved using the AdamW optimizer.

\subsection{Expert Network I: TransNeXt with Coordinate-Gated Mechanism}
\label{subsec:expert_transnext}

\begin{figure}[!t]
    \centering
    \includegraphics[width=0.95\linewidth]{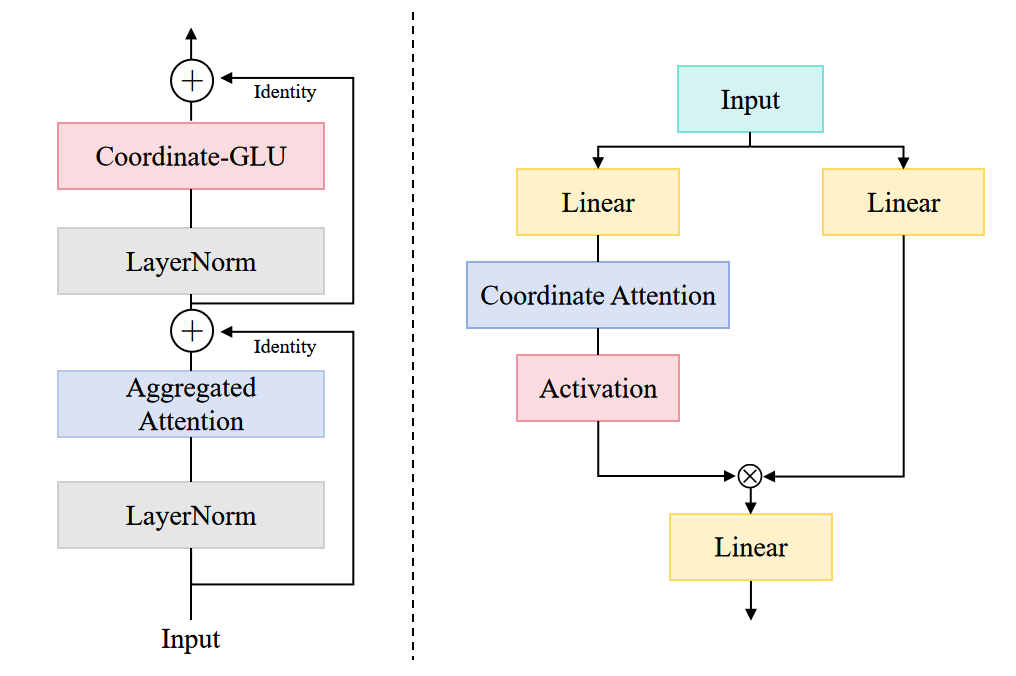}
    \caption{Architecture of the TransNeXt Expert Block. Left: The residual building block stacking the Coordinate-GLU and Aggregated Attention modules. Right: Detailed structure of the proposed Coordinate Attention Gated Linear Unit (CoordAttGLU), which replaces the standard Convolutional GLU to capture long-range dependencies along time and frequency axes.}
    \label{fig:co_glu_block}
\end{figure}

The first expert utilizes a customized TransNeXt backbone \cite{shi2024transnext}, specifically engineered to handle saturated jamming scenarios. In complex environments (e.g., triple-component mixtures), interference signals exhibit highly entangled features: global spectral occupancy (like the wideband sweep of LFM) coexists with fine-grained local textures (like the transient edges of Pulse jamming). To disentangle these features, this expert integrates Aggregated Attention for dual-scale perception and a CoordAttGLU for directional modeling. The structure of this expert is shown in Fig.~\ref{fig:co_glu_block}.

\subsubsection{Aggregated Attention for Saturated Interference}
To mitigate the feature collapse often observed in high-complexity jamming, we employ Aggregated Attention. Unlike standard self-attention, this mechanism balances global and local perception to match the physical characteristics of compound signals. The global path ($global$) captures long-range spectral dependencies—crucial for identifying continuous sweep trajectories across the entire bandwidth—while the local path ($\rho$) focuses on fine-grained textural details within sliding windows \cite{shi2024transnext}. Mathematically, for an input feature map $\mathbf{X}$, we first derive the query $\mathbf{Q}$, key $\mathbf{K}$, and value $\mathbf{V}$ through linear projections. Let $\hat{\mathbf{Q}}$, $\hat{\mathbf{K}}$, and $\hat{\mathbf{V}}$ denote their respective projected forms used in the attention mechanism. Furthermore, a relative position embedding $\mathbf{E}$ is incorporated into the query formulation to preserve spatial structure. The similarity terms $\mathbf{S}$ and the resulting aggregated attention matrix $\mathbf{A}_{(i,j)}$ are derived as follows:
\begin{align}
    \mathbf{S}^{\rho}_{(i,j)} &= (\hat{\mathbf{Q}}_{(i,j)} + \mathbf{QE}) \hat{\mathbf{K}}_{\rho(i,j)}^T, \label{eq:sim_terms_rho} \\
    \mathbf{S}^{global}_{(i,j)} &= (\hat{\mathbf{Q}}_{(i,j)} + \mathbf{QE}) \hat{\mathbf{K}}_{global(\mathbf{X})}^T, \label{eq:sim_terms_sigma} \\
    \begin{split}
        \mathbf{A}_{(i,j)} &= \text{Softmax}\bigg( \alpha \log N_{token} \cdot \text{Concat}\left(\mathbf{S}^{\rho}_{(i,j)}, \mathbf{S}^{global}_{(i,j)}\right) \\
        &\qquad + \mathbf{B}_{(i,j)} \bigg), 
    \end{split} \label{eq:agg_attn}
\end{align}
where $\mathbf{B}_{(i,j)}$ represents the learnable relative positional bias, $\alpha$ denotes the temperature parameter for the Softmax operation, and $N_{token}$ indicates the number of tokens involved in the attention calculation.

\subsubsection{Coordinate Attention Gated Linear Unit (CoordAttGLU)}
\label{subsubsec:coordattglu}
Jamming signals often possess distinct directional characteristics in the TF domain (e.g., STJ is a horizontal line, and Pulse is a vertical stripe). To capture these long-range dependencies along the time and frequency axes, we introduce the CoordAttGLU. Let $\mathbf{Z} \in \mathbb{R}^{H \times W \times C}$ be the input. The input tensor $\mathbf{Z}$ is projected and split along the channel dimension into two halves, $\mathbf{Z}_1, \mathbf{Z}_2 \in \mathbb{R}^{H \times W \times C/2}$. Coordinate Attention is applied to $\mathbf{Z}_1$ using two 1D global average pooling kernels, aggregating features into $\mathbf{z}_1^h$ and $\mathbf{z}_1^w$. These are concatenated and transformed to generate attention weights $\mathbf{g}^h$ and $\mathbf{g}^w$. The refined spatial feature map $\mathbf{Y}_{att}$ is obtained by re-weighting $\mathbf{Z}_1$, effectively highlighting the interference trajectory. Finally, the output is computed via element-wise multiplication with the gating branch $\mathbf{Z}_2$ given as follows:
\begin{align}
    \mathbf{z}_1^h(h) &= \frac{1}{W} \sum_{0 \le i < W} \mathbf{Z}_1(h, i), \label{eq:coord_pool_h} \\
    \mathbf{z}_1^w(y) &= \frac{1}{H} \sum_{0 \le j < H} \mathbf{Z}_1(j, y), \label{eq:coord_pool_w} \\
    [\mathbf{f}^h, \mathbf{f}^w] &= \text{Split}\left( \delta(F_{conv}([\mathbf{z}_1^h, \mathbf{z}_1^w])) \right), \label{eq:coord_split} \\
    \mathbf{g}^h &= \sigma(F_h(\mathbf{f}^h)), \quad \mathbf{g}^w = \sigma(F_w(\mathbf{f}^w)), \label{eq:coord_weights} \\
    \mathbf{Y}_{att}(h,y) &= \mathbf{Z}_1(h,y) \times \mathbf{g}^h(h) \times \mathbf{g}^w(y), \label{eq:coord_apply} \\
    \mathbf{Y}_{out} &= \text{Linear}_{out}\left( \text{GELU}(\mathbf{Y}_{att}) \odot \mathbf{Z}_2 \right), \label{eq:coord_out}
\end{align}
where $\delta$ denotes a non-linear activation function, $\sigma$ represents the Sigmoid function, and $\odot$ denotes element-wise multiplication (i.e., the Hadamard product).

\subsection{Expert Network II: Selective Kernel GhostNet}
\label{subsec:expert_skghost}

The second expert targets the challenge of bandwidth variance. GNSS interference signals exhibit drastic variations in spectral occupancy—ranging from extremely narrowband STJ to wideband PBNJ or LFM. Fixed-kernel convolutions struggle to capture such multi-scale patterns efficiently. To address this, we introduce the \textit{Selective Kernel GhostNet} (SK-GhostNet), which integrates Ghost Modules \cite{han2020ghostnet} with a dynamic SK mechanism \cite{li2019selective}.

\begin{figure}[!t]
    \centering
    \includegraphics[width=0.8\linewidth]{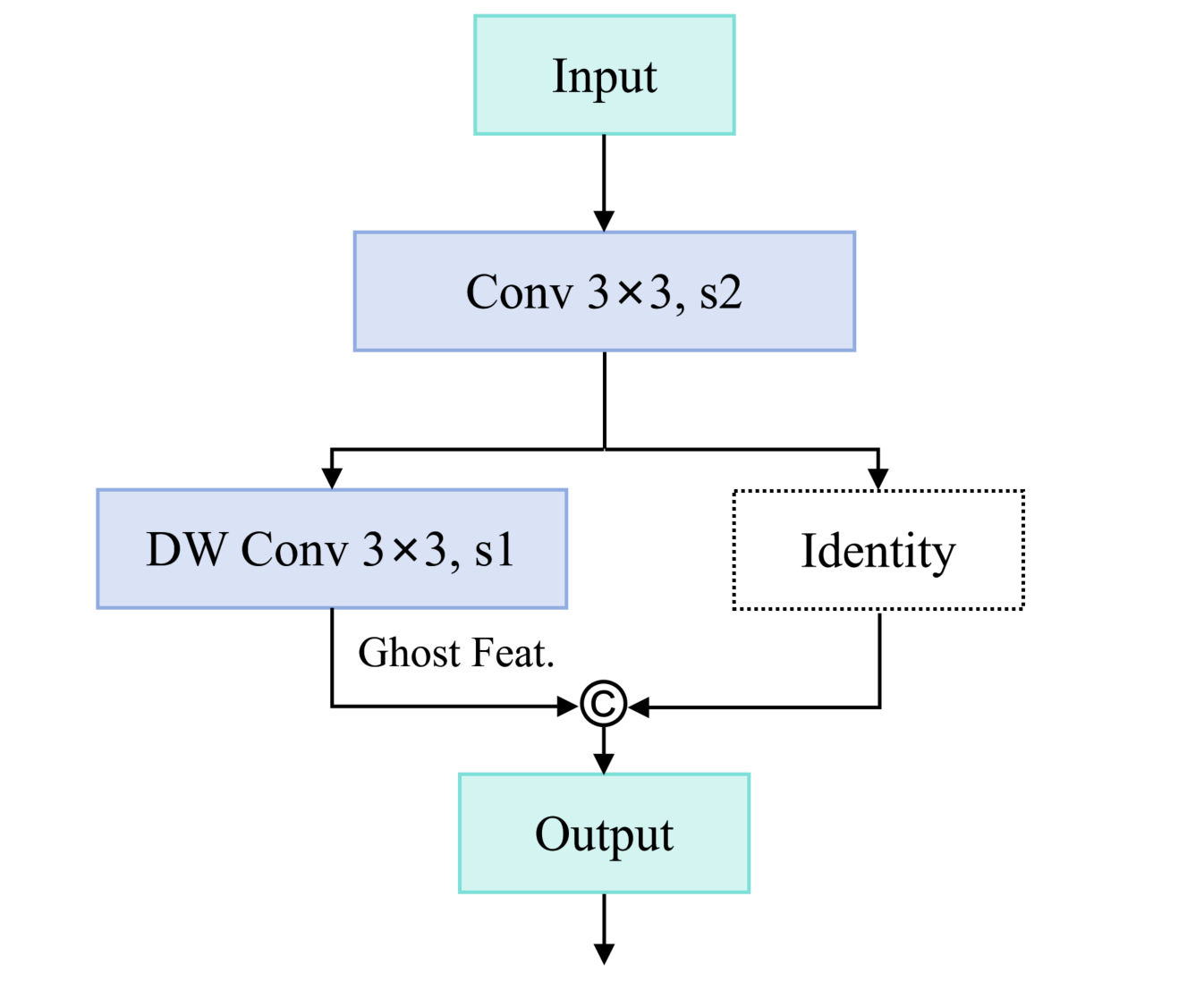}
    \caption{The schematic of the Ghost Module used as the fundamental building block in Expert II. It divides the feature extraction process into a primary convolution branch and a cheap linear operation branch (Depthwise Convolution), effectively reducing parameter redundancy.}
    \label{fig:ghost_module}
\end{figure}

\subsubsection{The Ghost Module Strategy}
To maintain efficiency, the Ghost Module (shown in Fig.~\ref{fig:ghost_module}) divides feature extraction into a primary convolution generating intrinsic maps $\mathbf{Y}'$ and a series of cheap linear operations $\Phi$ generating "ghost" features $\mathbf{Y}''$ given as follows \cite{han2020ghostnet}:
\begin{align}
    \mathbf{Y}' &= \mathbf{X} * \mathbf{W}_{primary}, \label{eq:ghost_primary} \\
    \mathbf{Y}_{ghost} &= \text{Concat}(\mathbf{Y}', \Phi(\mathbf{Y}')), \label{eq:ghost_concat}
\end{align}
where $\mathbf{W}_{primary}$ represents the primary convolutional kernels.

\subsubsection{Adaptive SK-Ghost Block for Multi-Scale Features}
The core contribution of Expert II is the SK-Ghost Block, which employs a Split-Fuse-Select strategy to adaptively adjust the receptive field size \cite{li2019selective}. This allows the network to focus on localized frequency spikes for narrowband jammers or aggregate global context for wideband noise. In the split phase, the input is processed by parallel Ghost branches with distinct kernel sizes to extract multi-scale features $\mathbf{U}_q$. These are aggregated via element-wise summation to form $\mathbf{U} = \sum_{q} \mathbf{U}_q$. A global descriptor $\mathbf{z}$ is then generated via global average pooling over $\mathbf{U}$. The attention scores $\mathbf{a}_{q,c}$ and the final output are computed as follows:
\begin{align}
    \mathbf{U}_q &= \mathcal{T}_q(\mathbf{X}), \quad q=1, \dots, Q, \label{eq:sk_split} \\
    \mathbf{a}_{q,c} &= \frac{e^{\mathbf{W}_q \mathbf{z}}}{\sum_{j=1}^{Q} e^{\mathbf{W}_j \mathbf{z}}}, \label{eq:sk_softmax} \\
    \mathbf{V}_c &= \sum_{q=1}^{Q} \mathbf{a}_{q,c} \cdot \mathbf{U}_{q,c}, \label{eq:sk_aggregate}
\end{align}
where $\mathbf{W}_q$ denotes the trainable weights of the fully connected layer for the $q$-th branch. This dynamic aggregation ensures scale-invariance, enabling robust classification across diverse jamming bandwidths \cite{yang2023complexsk}.

\subsection{Expert Network III: MobileNetV4 with Universal Inverted Bottlenecks}
\label{subsec:expert_mnv4}

To ensure the framework remains universally efficient, particularly for processing simple jamming primitives (e.g., single-tone or pulse), the third expert adopts the \textit{MobileNetV4} (MNv4) architecture \cite{qin2024mobilenetv4}. This expert is designed to achieve Pareto-optimal performance by utilizing Universal Inverted Bottlenecks (UIB) and Mobile Multi-Query Attention (Mobile MQA).

\subsubsection{Universal Inverted Bottleneck (UIB)}
The UIB unifies efficient micro-architectures by introducing optional depthwise convolutions. This flexibility allows Expert III to decouple the receptive field size from network depth, ensuring robust feature extraction for fundamental signal types with minimal FLOPs \cite{qin2024mobilenetv4}.

\subsubsection{Mobile Multi-Query Attention (Mobile MQA)}
To model global contexts without the high cost of standard Self-Attention, Mobile MQA shares Key and Value heads and employs Spatial Reduction (SR) via stride-2 depthwise convolutions. The attention operation, normalized by the scaling factor $\sqrt{d_k}$ where $d_k$ is the dimension of the key head, is defined as:

\begin{equation}
\begin{split}
    \text{Attention}(\mathbf{Q}, \mathbf{K}, \mathbf{V}) = & \text{Softmax}\left( \frac{\mathbf{Q} (\text{SR}(\mathbf{X})\mathbf{W}_K)^T}{\sqrt{d_k}} \right) \\
    & \times (\text{SR}(\mathbf{X})\mathbf{W}_V), 
\end{split}
\label{eq:mobile_mqa}
\end{equation}
where $\mathbf{W}_K$ and $\mathbf{W}_V$ are the learnable projection matrices for Keys and Values, respectively. This design provides a lightweight mechanism to handle low-entropy signals where complex feature disentanglement is unnecessary.

\subsection{Physics-Guided Gating Mechanism}
\label{subsec:gating_mechanism}

To achieve an effective trade-off between classification precision and computational efficiency, the proposed framework incorporates a physics-guided gating mechanism. Unlike static neural networks, our router dynamically allocates computational resources by implicitly learning the spectral complexity of the input signal. Let $\mathbf{X}_{\text{PSD}} \in \mathbb{R}^{H \times W}$ denote the input PSD image, which serves as a high-fidelity descriptor of the physical characteristics of the interference \cite{wang2017statistical}.

To minimize computational overhead, we employ a \textit{feature reuse strategy}. The PSD input is first processed by a lightweight convolutional encoder consisting of two stacked $3\times 3$ Conv-Pooling blocks to extract spectral feature maps. These spectral features serve a dual purpose: they are simultaneously fused into the main expert streams via the Squeeze-and-Excitation Fusion module to enhance feature selectivity and fed into the router for expert selection \cite{xiao2025compound}. Consequently, the gating network $G(\cdot)$ effectively requires only a lightweight linear projection head on top of these shared features. This design implies that the physics-guided decision-making process incurs negligible marginal cost, as the spectral feature extraction is a prerequisite for the fusion task.

Based on these shared spectral features, the router learns a non-linear mapping between the visual topology of the PSD and the required model capacity. It is worth noting that the router does not explicitly calculate a scalar entropy value as a threshold. Instead, it categorizes the jamming environment into three physical complexity levels. Specifically, in scenarios characterized by high SNR or simple jamming primitives where the PSD exhibits sparse and distinct peaks, the router tends to activate the lightweight MobileNetV4, thereby reducing FLOPs and latency for trivial samples. As the spectral energy becomes distributed over a wider yet structured range, typical of signals with varying bandwidths or moderate non-stationary characteristics, the system generally selects the SK-GhostNet. This expert efficiently handles multi-scale features via SK without resorting to maximum computational overhead. Finally, in saturated environments or low-SNR scenarios where the spectral structure becomes chaotic and highly entangled, the router invokes the higher-capacity TransNeXt \cite{shi2024transnext}. The utilization of global receptive fields and coordinate attention is employed here to disentangle coupled features and maintain recognition robustness. This hierarchical allocation strategy enables the receiver to operate as a ``Green'' cognitive agent, maintaining an energy-efficient state during nominal conditions while scaling up computational resources when necessitated by sophisticated interference environments \cite{chen2024cognitive}.

\section{Simulation Results and Analysis}
\label{sec:simulation}

\subsection{Dataset Generation and Preprocessing}
To rigorously evaluate the proposed classification framework, we established a large-scale, high-fidelity dataset covering single-source and complex compound jamming scenarios using a discrete-time baseband model in MATLAB, adhering to the standard signal reception formulation described in \cite{wang2018detection}. The simulation environment is configured with a sampling frequency of $20$ MHz, where each signal sample persists for $1$ ms, yielding a sequence length of $N = 20,000$ samples. These signals are transmitted through an AWGN channel, with the JNR systematically swept from $-25$ dB to $15$ dB in $1$ dB increments. This wide dynamic range is selected to encompass both the ``submerged'' scenarios typical of early-stage interference and high-power saturation cases, consistent with recent benchmarks in the literature \cite{mehr2025deep, jia2025lightweight}.

The dataset comprises 21 distinct jamming categories structured into three complexity tiers: five fundamental single-source types (STJ, MTJ, LFM, Pulse, and PBNJ); nine dual-component combinations generated via linear superposition, where a PR is uniformly randomized between $-3$ dB and $3$ dB to simulate realistic power disparities \cite{xiao2025compound}; and seven multi-component scenarios involving the simultaneous presence of three jamming types. To maintain consistency with the dual-component scenarios, strictly equal power allocation is avoided; instead, the power weights of the three components are randomized such that the maximum relative power difference between any two components remains within the $[-3, 3]$ dB range. This setup creates a highly dynamic saturated interference environment challenging for static classifiers.

Following time-domain synthesis, the raw I/Q data is transformed into dual-domain image representations—specifically Time-Frequency Images (TFI) and PSD images—all standardized to a resolution of $224 \times 224$ pixels. The TFI is constructed using the STFT with a 128-point periodic Hann window and a high overlap of approximately $92\%$ ($N_{over} \approx 117$). This high overlap ratio is chosen to ensure fine temporal resolution and continuity for non-stationary signals like chirp jamming \cite{cai2019waterfall}. The spectrograms are processed with a 4096-point FFT, followed by Gamma correction ($\gamma=0.9$) and logarithmic mapping to enhance the visibility of low-power components against the noise floor. Complementing the TFI, the PSD image is generated using the MTM with a time-bandwidth product of $NW=3$ and $K=5$ orthogonal Slepian tapers. Unlike standard periodograms, MTM is employed here to ensure variance-reduced spectral estimation, which is critical for distinguishing the spectral shape of overlapping compound components \cite{xiao2025compound, wang2017statistical}. These spectral curves are plotted in a logarithmic scale without axes, grids, or borders to compel the neural network to focus exclusively on spectral shape features. 

To strictly ensure the statistical diversity of the dataset and mitigate the risk of overfitting to trivial patterns, we enforced rigorous parameter randomization across all dimensions (e.g., carrier frequency, initial phase, and duty cycle). This strategy breaks the temporal correlation between samples, effectively simulating the non-IID characteristics of real-world non-stationary channels. By maximizing the distributional variance within each class, we implicitly enlarge the Wasserstein distance between sample clusters, ensuring that the model learns robust, generalized physical features rather than memorizing specific synthesis artifacts.

The data generation process was parallelized to produce 1,000 independent Monte Carlo samples for each of the 21 classes at every JNR point, resulting in a total dataset size of 861,000 samples, which ensures the model is trained on a statistically significant distribution of signal variations.

\subsection{Implementation Details}
\label{subsec:implementation}

\subsubsection{Hardware and Software Environment}
Experiments were conducted on a workstation equipped with an Intel Core i9-12900KF CPU, an NVIDIA GeForce RTX 5060 Ti GPU (16 GB), and 64 GB of RAM. The proposed model was implemented using PyTorch 2.x and Python 3.11, accelerated by CUDA 11.8 and cuDNN.

\subsubsection{Training Strategy}
To achieve optimal convergence and generalization, the network training was governed by a rigorous optimization strategy. We employed the AdamW optimizer, which decouples weight decay from gradient updates, with an initial learning rate set to $\eta = 1 \times 10^{-4}$ and a weight decay coefficient of 0.05. The learning rate followed a hybrid scheduling policy: a linear warm-up phase was applied for the first 10 epochs to stabilize the initial gradients. This is particularly important for MoE architectures to prevent early gating instability where the router might collapse to a single expert before features are learned \cite{lepikhin2020gshard, shi2024transnext}. This was followed by a Cosine Annealing decay strategy for the remaining epochs to facilitate convergence into a flat local minimum. The maximum number of training epochs was set to 50, with a batch size of 16.

To mitigate the risk of overfitting, particularly given the complex feature space of compound jamming signals, we introduced multiple regularization techniques. First, \textit{DropPath} (Stochastic Depth) was applied with a rate of 0.2 during the backbone feature extraction. This stochastic dropping of layers forces the network to learn robust features across different depths and is essential for deep architectures such as TransNeXt \cite{shi2024transnext}. Second, an \textit{Early Stopping} mechanism was implemented, halting the training process if the validation accuracy did not improve for 15 consecutive epochs (patience). Furthermore, data augmentation techniques, including random horizontal flips and affine translations, were applied to the training set to enhance the model's robustness against signal variations \cite{ferre2019jammer}. The dataset was randomly partitioned with a ratio of 80\% for training and 20\% for validation.

\subsubsection{Signal Parameterization}
The synthetic dataset was constructed based on the kinematic and spectral characteristics of five fundamental jamming primitives: STJ, MTJ, LFM, Pulse, and PBNJ. To simulate realistic electromagnetic environments, the signals were generated at a sampling rate of $F_s = 20$ MHz with a duration of 1 ms. Key parameters for each signal type—such as carrier frequency, duty cycle, sweep period, and bandwidth—were randomized within physically meaningful ranges to ensure diversity.

Detailed parameter configurations for the jamming scenarios are summarized in Table \ref{tab:jamming_params}. These primitives were further combined to form hierarchical classes, including single, dual, and multi-component interference signals. For compound signals (Dual and Multi-component), the PR between different components was randomized within $\pm 3$ dB to simulate varying signal strengths, and the JNR ranged from -25 dB to 15 dB.

\begin{table}[htbp]
\caption{Parameter Settings for Generated Jamming Signals}
\label{tab:jamming_params}
\centering
\renewcommand{\arraystretch}{1.2} 
\begin{tabular}{llc} 
\toprule
\textbf{Signal Type} & \textbf{Parameter} & \textbf{Value / Range} \\
\midrule
\multirow{2}{*}{Common} & Sampling Rate ($F_s$) & 20 MHz \\
 & Signal Duration & 1 ms \\
\midrule 
\multirow{2}{*}{STJ} & Carrier Frequency ($f_c$) & $\pm 9.5$ MHz \\
 & Initial Phase & $0 \sim 2\pi$ \\
\midrule 
\multirow{3}{*}{MTJ} & Number of Tones & $5 \sim 7$ \\
 & Tone Spacing & $1.5 \sim 3.0$ MHz \\
 & Frequency Range & $\pm 9.5$ MHz \\
\midrule 
\multirow{2}{*}{LFM} & Sweep Bandwidth & 10 MHz \\
 & Sweep Period & 1 ms \\
\midrule 
\multirow{3}{*}{Pulse} & Duty Cycle & $\approx 30\%$ \\
 & Pulse Repetition Interval (PRI) & $1/6$ ms \\
 & Carrier Frequency & $\pm 9.5$ MHz \\
\midrule 
\multirow{2}{*}{PBNJ} & Noise Bandwidth & $10\% \sim 25\% F_s$ \\
 & Center Frequency & Randomized \\
\midrule 
\multirow{2}{*}{Compound Settings} & JNR Range & $-25 \sim 15$ dB \\
 & Power Ratio (PR) & $\pm 3$ dB \\
\bottomrule
\end{tabular}
\end{table}

\begin{table}[htbp]
    \centering
    \caption{OA and Complexity of Various Methods}
    \label{tab:comparison}
   
    \resizebox{\columnwidth}{!}{%
        \begin{tabular}{lcccc}
            \toprule
            Methods & OA(\%) & Active Params.(M) & FLOPs(G) & Time(ms) \\
            \midrule
            ViT-B/16        & 86.76 & 85.80 & 11.29 & 6.56  \\
            RCNN            & 89.93 & 13.12 & 0.55  & 0.38  \\
            ResNet-18       & 92.44 & 11.18 & 1.82  & 1.14  \\
            EfficientNet-B0 & 93.41 & 4.02  & 0.41  & 3.76  \\
            TFPENet         & 95.48 & 11.57 & 2.56  & 3.45  \\
            \textbf{PhyG-MoE (Ours)} & \textbf{97.58} & \textbf{11.19} & \textbf{1.71} & \textbf{4.08} \\
            \bottomrule
        \end{tabular}%
    }
\end{table}

\begin{table*}[htbp]
\centering
\renewcommand{\arraystretch}{1.25}
\caption{Comprehensive Precision(\%), Recall(\%), and F1 Score(\%) of Various Recognition Methods}
\label{tab:combined_final_correct_bold}

\resizebox{\textwidth}{!}{%
\begin{tabular}{c|ccc|ccc|ccc|ccc|ccc|ccc|ccc}
\toprule
\multirow{2}{*}{Methods} & \multicolumn{3}{c|}{LFM} & \multicolumn{3}{c|}{MTJ} & \multicolumn{3}{c|}{PBNJ} & \multicolumn{3}{c|}{Pulse} & \multicolumn{3}{c|}{STJ} & \multicolumn{3}{c|}{LFM+PBNJ} & \multicolumn{3}{c}{LFM+Pulse} \\
\cline{2-22} 
 & Pr & R & F1 & Pr & R & F1 & Pr & R & F1 & Pr & R & F1 & Pr & R & F1 & Pr & R & F1 & Pr & R & F1 \\
\hline
R-CNN & \underline{95.23} & 87.17 & 91.02 & 79.16 & 85.87 & 82.38 & 83.27 & 81.99 & 82.63 & 91.85 & 94.06 & 92.94 & 89.10 & 89.46 & 89.28 & 75.44 & 76.43 & 75.93 & 86.88 & 78.02 & 82.21 \\
ResNet-18 & 94.67 & \underline{90.58} & 92.58 & 85.71 & 89.25 & 87.44 & 83.45 & 89.86 & 86.54 & 97.29 & 95.21 & 96.24 & 96.86 & 91.74 & 94.23 & \underline{90.37} & 79.18 & 84.40 & \underline{94.60} & 82.07 & 87.89 \\
ViT-B/16 & 85.03 & 85.35 & 85.19 & 75.00 & 83.72 & 79.12 & 81.62 & 81.02 & 81.32 & 94.68 & 89.20 & 91.86 & 90.68 & 85.49 & 88.01 & 83.93 & 73.04 & 78.11 & 93.40 & 77.96 & 84.99 \\
EfficientNet & 94.21 & 89.10 & 91.58 & 78.40 & 88.50 & 83.15 & \underline{87.50} & 84.76 & 86.11 & 96.34 & 92.34 & 94.30 & 89.69 & 90.46 & 90.07 & 87.88 & 80.20 & 83.87 & 93.05 & 81.71 & 87.01 \\
TFPENet & 94.62 & 89.41 & \underline{93.79} & \underline{95.69} & \underline{93.05} & \underline{95.31} & 85.05 & \textbf{98.52} & \underline{91.29} & \underline{98.13} & \underline{96.81} & \underline{97.46} & \underline{98.84} & \underline{99.06} & \underline{98.95} & 87.03 & \underline{90.46} & \underline{88.71} & 94.22 & \underline{84.45} & \underline{89.07} \\
PhyG-MoE & \textbf{95.45} & \textbf{93.80} & \textbf{94.62} & \textbf{97.55} & \textbf{96.40} & \textbf{96.97} & \textbf{93.20} & \underline{95.15} & \textbf{94.16} & \textbf{99.10} & \textbf{98.55} & \textbf{98.82} & \textbf{99.60} & \textbf{99.72} & \textbf{99.66} & \textbf{92.50} & \textbf{96.85} & \textbf{89.10} & \textbf{94.85} & \textbf{85.60} & \textbf{90.01} \\
\bottomrule
\end{tabular}%
}

\vspace{5pt}

\resizebox{\textwidth}{!}{%
\begin{tabular}{c|ccc|ccc|ccc|ccc|ccc|ccc|ccc}
\toprule
\multirow{2}{*}{Methods} & \multicolumn{3}{c|}{MTJ+LFM} & \multicolumn{3}{c|}{MTJ+PBNJ} & \multicolumn{3}{c|}{MTJ+Pulse} & \multicolumn{3}{c|}{Pulse+PBNJ} & \multicolumn{3}{c|}{STJ+LFM} & \multicolumn{3}{c|}{STJ+PBNJ} & \multicolumn{3}{c}{STJ+Pulse} \\
\cline{2-22} 
 & Pr & R & F1 & Pr & R & F1 & Pr & R & F1 & Pr & R & F1 & Pr & R & F1 & Pr & R & F1 & Pr & R & F1 \\
\hline
R-CNN & 71.56 & 76.87 & 74.12 & 70.74 & 79.65 & 74.94 & 74.97 & 77.75 & 76.33 & 75.52 & 77.11 & 76.31 & 84.98 & 79.38 & 82.08 & 80.02 & 78.91 & 79.46 & 85.43 & 77.58 & 81.32 \\
ResNet-18 & 84.86 & 81.71 & 83.26 & 67.99 & 87.06 & 76.36 & 77.19 & 85.32 & 81.05 & 89.21 & 81.02 & 84.92 & 83.54 & 81.71 & 87.23 & 76.49 & 86.25 & 81.08 & 85.97 & \underline{85.97} & 85.97 \\
ViT-B/16 & 69.70 & 78.90 & 74.02 & 61.43 & 78.66 & 68.98 & 63.87 & 83.08 & 72.22 & \textbf{90.95} & 70.70 & 79.56 & 86.37 & 77.17 & 81.51 & 77.90 & 81.01 & 79.43 & 91.32 & 79.16 & 84.81 \\
EfficientNet & 70.65 & \underline{83.40} & 76.50 & 63.86 & 86.06 & 73.32 & 83.95 & 80.95 & 82.42 & 85.44 & 80.98 & 83.15 & \underline{92.59} & 81.42 & 86.65 & \underline{87.96} & 82.11 & \underline{84.93} & 87.95 & 83.82 & 85.84 \\
TFPENet & \underline{93.57} & 82.39 & \underline{87.63} & \underline{72.49} & \textbf{92.08} & \underline{81.12} & \underline{94.12} & \underline{86.39} & \underline{91.00} & \underline{89.45} & \underline{87.45} & \underline{88.44} & 92.24 & \underline{86.26} & \underline{89.15} & 76.20 & \underline{93.77} & 84.08 & \underline{94.00} & 85.31 & \underline{90.78} \\
PhyG-MoE & \textbf{95.60} & \textbf{85.55} & \textbf{90.30} & \textbf{90.15} & \underline{88.40} & \textbf{89.27} & \textbf{95.25} & \textbf{89.60} & \textbf{92.34} & 88.75 & \textbf{90.30} & \textbf{89.52} & \textbf{93.54} & \textbf{89.85} & \textbf{91.52} & \textbf{89.45} & \textbf{94.10} & \textbf{91.72} & \textbf{96.20} & \textbf{94.85} & \textbf{95.52} \\
\bottomrule
\end{tabular}%
}

\vspace{5pt}

\resizebox{\textwidth}{!}{%
\begin{tabular}{c|ccc|ccc|ccc|ccc|ccc|ccc|ccc}
\toprule
\multirow{2}{*}{Methods} & \multicolumn{3}{c|}{LFM+Pls+PBNJ} & \multicolumn{3}{c|}{MTJ+LFM+PBNJ} & \multicolumn{3}{c|}{MTJ+LFM+Pls} & \multicolumn{3}{c|}{MTJ+Pls+PBNJ} & \multicolumn{3}{c|}{STJ+LFM+PBNJ} & \multicolumn{3}{c|}{STJ+LFM+Pls} & \multicolumn{3}{c}{STJ+Pls+PBNJ} \\
\cline{2-22} 
 & Pr & R & F1 & Pr & R & F1 & Pr & R & F1 & Pr & R & F1 & Pr & R & F1 & Pr & R & F1 & Pr & R & F1 \\
\hline
R-CNN & 74.51 & 74.81 & 74.66 & 83.02 & 80.17 & 81.57 & 78.52 & 72.87 & 75.59 & 84.36 & 77.38 & 80.72 & 78.37 & 82.52 & 80.39 & 68.07 & 82.14 & 74.45 & 82.61 & 76.22 & 79.29 \\
ResNet-18 & 77.69 & 77.95 & 77.81 & \underline{85.50} & 81.24 & \underline{83.31} & 79.84 & 77.38 & 78.59 & 68.00 & \textbf{85.86} & 75.89 & 81.62 & 80.78 & 85.86 & 88.20 & 81.38 & 84.65 & 81.32 & \underline{82.52} & 81.92 \\
ViT-B/16 & 63.30 & 74.62 & 68.50 & 77.88 & 76.29 & 77.08 & 76.82 & 71.33 & 73.97 & 84.74 & 74.62 & 79.36 & 74.22 & 80.29 & 77.13 & 86.88 & 72.11 & 78.81 & 69.14 & 77.93 & 73.27 \\
EfficientNet & 79.67 & 78.22 & 78.94 & 71.12 & 85.49 & 77.64 & \underline{88.49} & 73.51 & 80.31 & 71.43 & 83.50 & 76.99 & \underline{85.77} & 80.01 & 82.79 & \underline{91.69} & 78.07 & \underline{85.58} & 78.34 & 82.20 & 80.22 \\
TFPENet & \underline{85.91} & \underline{84.19} & \underline{85.04} & 70.80 & \underline{89.72} & 79.15 & 87.63 & \underline{81.21} & \underline{84.30} & \underline{91.66} & 80.76 & \underline{85.87} & 83.33 & \underline{91.16} & \underline{87.07} & 86.31 & \underline{83.71} & 84.99 & \underline{92.47} & 82.10 & \underline{86.98} \\
PhyG-MoE & \textbf{94.10} & \textbf{88.35} & \textbf{91.13} & \textbf{88.50} & \textbf{95.40} & \textbf{86.13} & \textbf{95.85} & \textbf{83.75} & \textbf{89.39} & \textbf{95.20} & \underline{85.10} & \textbf{89.87} & \textbf{91.62} & \textbf{94.45} & \textbf{91.59} & \textbf{92.65} & \textbf{85.80} & \textbf{89.09} & \textbf{93.65} & \textbf{86.20} & \textbf{89.77} \\
\bottomrule
\end{tabular}%
}
\end{table*}

\subsection{Performance Evaluation Metrics}
\label{subsec:evaluation_metrics}

To provide a comprehensive assessment of the proposed model's classification capabilities and computational efficiency, we utilized standard quantitative metrics \cite{ferre2019jammer, vandermerwe2024optimal}. These include Overall Accuracy (OA), Precision, Recall, F1-Score, and computational complexity indicators.

\subsubsection{Accuracy Metrics}
The primary indicator of global model performance is OA, which represents the proportion of correctly identified jamming samples relative to the total test set size. Let $N_{total}$ denote the total number of query samples and $N_{correct}$ represent the count of accurate predictions. OA is formulated as \cite{mehr2025deep}:
\begin{equation}
\text{OA} = \frac{N_{correct}}{N_{total}} \times 100\%.
\label{eq:oa}
\end{equation}

\subsubsection{Precision and Recall}
Given the potential imbalance in compound jamming scenarios, class-wise performance is evaluated using Precision and Recall \cite{liu2024gnss}. Precision ($\mathcal{P}_c$) quantifies the reliability of the model when it predicts a specific class $c$, whereas Recall ($\mathcal{R}_c$) measures the model's ability to retrieve all instances of that class. These are defined based on True Positives ($TP_c$), False Positives ($FP_c$), and False Negatives ($FN_c$) given as follows:
\begin{equation}
\mathcal{P}_c = \frac{TP_c}{TP_c + FP_c}, \quad \mathcal{R}_c = \frac{TP_c}{TP_c + FN_c}.
\label{eq:prec_rec}
\end{equation}

\subsubsection{F1-Score}
To balance the trade-off between Precision and Recall, the F1-Score is computed. It serves as the harmonic mean of the two metrics, providing a single scalar value that reflects the robustness of the classifier, especially for difficult-to-distinguish jamming classes given as follows \cite{vandermerwe2024optimal}:
\begin{equation}
F1_c = 2 \times \frac{\mathcal{P}_c \times \mathcal{R}_c}{\mathcal{P}_c + \mathcal{R}_c}.
\label{eq:f1}
\end{equation}

\subsubsection{Confusion Matrix}
To visualize the misclassification behavior, we employ the Confusion Matrix, denoted as $\mathbf{C} \in \mathbb{R}^{N_{class} \times N_{class}}$, where $N_{class}$ is the number of jamming categories. Each entry $C_{i,j}$ indicates the number of samples with ground truth label $i$ that were classified as label $j$ \cite{ferre2019jammer}. This matrix is instrumental in identifying specific confusion patterns between spectrally similar jamming types.

\subsubsection{Computational Complexity}
Beyond classification accuracy, the theoretical computational cost is evaluated using FLOPs. This metric aggregates the number of multiply-accumulate (MAC) operations required for a single forward pass \cite{jia2025lightweight}. For convolutional layers, the complexity is derived from the kernel size $k_{size} \times k_{size}$, input/output channels ($C_{in}, C_{out}$), and the spatial dimensions of the feature map ($H \times W$) given as follows:
\begin{equation}
\text{FLOPs}_{conv} = \sum_{l=1}^{L} 2 \cdot H_l W_l \cdot (k_{size})_l^2\cdot C_{in}^{(l)} \cdot C_{out}^{(l)},
\label{eq:flops}
\end{equation}
where $L$ is the number of convolutional layers. Similarly, for the linear layers in the projection heads, the complexity is proportional to the input and output neuron counts. This hardware-independent metric allows for a fair comparison of the proposed architecture against existing state-of-the-art methods \cite{han2020ghostnet, jia2025lowpower}.

\subsection{Performance Analysis}

\subsubsection{Overall Comparison and Computational Efficiency}
The quantitative comparison of the proposed PhyG-MoE framework against state-of-the-art baselines is presented in Table \ref{tab:comparison}. In terms of recognition capability, PhyG-MoE achieves a superior OA of 97.58\%, outperforming the recent dual-stream TFPENet by 2.1\% and the standard ResNet-18 by 5.14\%. This performance gain is attributed to the heterogeneous expert pool, which effectively captures both local spectral textures and global temporal dependencies.

Critically, this accuracy does not come at the expense of computational exorbitance. As evidenced in Table \ref{tab:comparison}, while the \textit{Total Parameter Capacity} of our mixture model is 32.27 M, the conditional execution strategy ensures that the \textit{Active Parameters} per inference are only 11.19 M. Consequently, PhyG-MoE requires only 1.71 GFLOPs, which is 33.2\% lower than TFPENet (2.56 GFLOPs) and comparable to the lightweight ResNet-18 (1.82 GFLOPs). It is important to note that the proposed dynamic routing does not introduce a computational bottleneck. Since the PSD feature encoder is shared with the Squeeze-and-Excitation (SE) Fusion module, the marginal cost of the routing operation is strictly limited to the final classification head (a single MLP). Theoretical analysis shows that the specific overhead of the router accounts for less than 0.5\% of the total inference FLOPs. This result validates that the proposed framework successfully breaks the static trade-off between model depth and inference cost, making it highly suitable for energy-constrained SAGIN edge nodes.

\subsubsection{Robustness Across JNR Regimes}
\begin{figure}[t]
    \centering
    \includegraphics[width=0.9\linewidth]{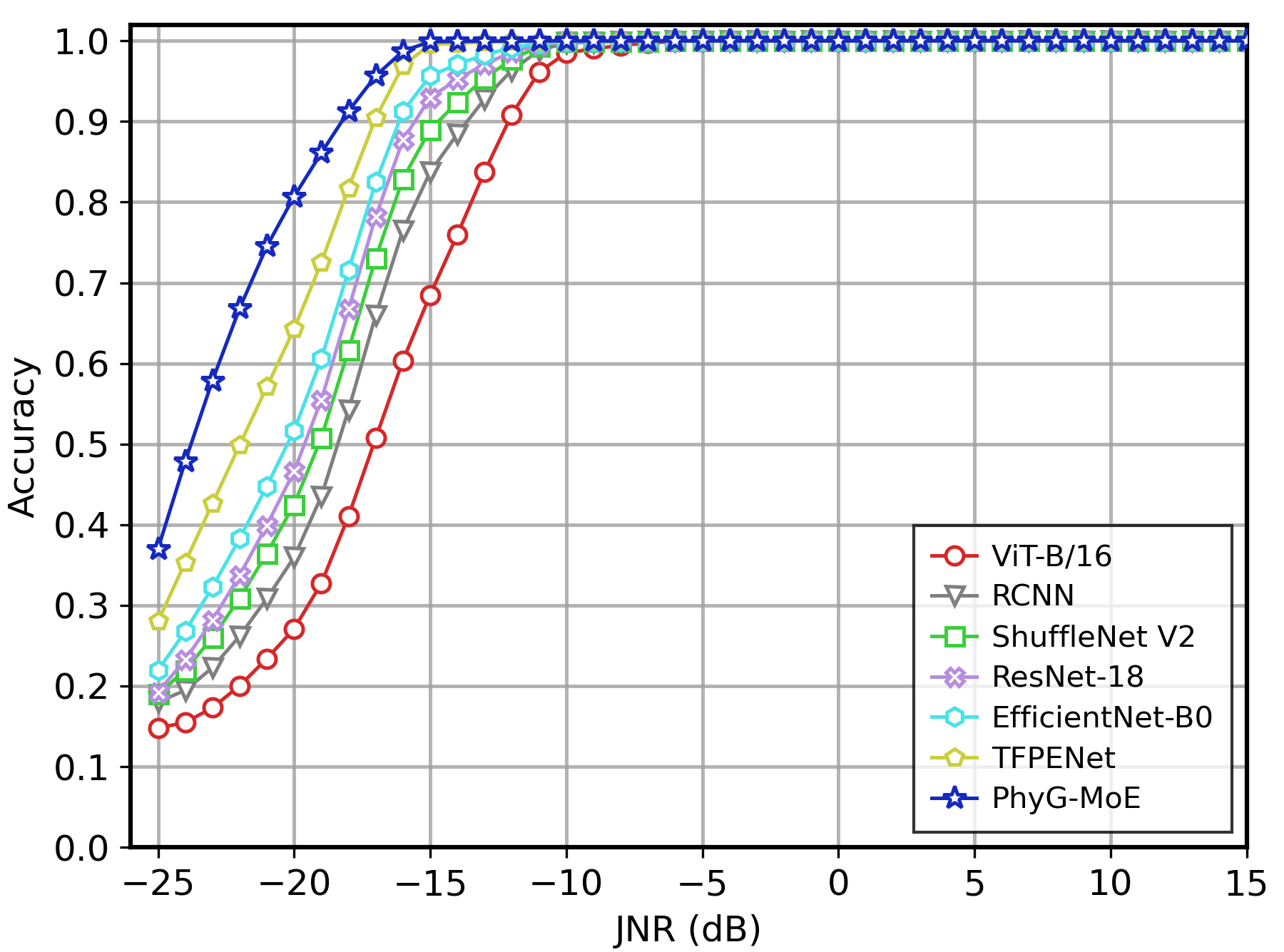}
    \caption{Classification accuracy versus JNR. The proposed PhyG-MoE demonstrates superior robustness, achieving rapid convergence to high accuracy even in low-SNR regimes.}
    \label{fig:acc_jnr}
\end{figure}

To evaluate the model's reliability under varying channel conditions, Fig. \ref{fig:acc_jnr} (referencing the uploaded AC.png) illustrates the classification accuracy as a function of JNR ranging from -25 dB to 15 dB.

It is observed that the proposed PhyG-MoE (blue star curve) exhibits the most robust noise immunity. Specifically, PhyG-MoE reaches the 90\% accuracy threshold at approximately -18 dB, whereas ResNet-18 and EfficientNet-B0 require -14 dB and -12 dB, respectively, to achieve comparable performance. In the extremely low JNR regime (e.g., -24 dB), where signal features are heavily submerged in noise, our method maintains an accuracy above 45\%, significantly surpassing ViT-B/16 ($\approx$ 15\%). This robustness is largely credited to the MTM-based spectral input and the TransNeXt expert's ability to extract coherent features from low-SNR environments via the coordinate-gated mechanism.

\label{subsec:performance_analysis}

\subsubsection{Inter-Class Discriminability Analysis}
The class-wise performance reported in Table \ref{tab:combined_final_correct_bold} highlights the advantage of our approach in handling compound interference. For fundamental primitives, most models achieve high F1-scores ($>$90\%). However, significant performance degradation is observed in baselines when processing Triple-Component signals.

For instance, in the complex \textit{STJ+LFM+PBNJ} scenario, TFPENet and ResNet-18 achieve F1-scores of 87.07\% and 85.86\%, respectively. In contrast, PhyG-MoE maintains a high F1-score of 91.59\%. Similarly, for \textit{LFM+Pulse+PBNJ}, which represents the most dynamic interference type, our model improves the Recall by over 10\% compared to ResNet-18. This indicates that the specialized experts are capable of disentangling the mutually masking features of wideband noise and sweep signals, minimizing confusion between spectrally similar categories.

\subsubsection{Interpretability of Physics-Guided Routing}
\begin{figure}[!t]
    \centering
    \includegraphics[width=0.9\linewidth]{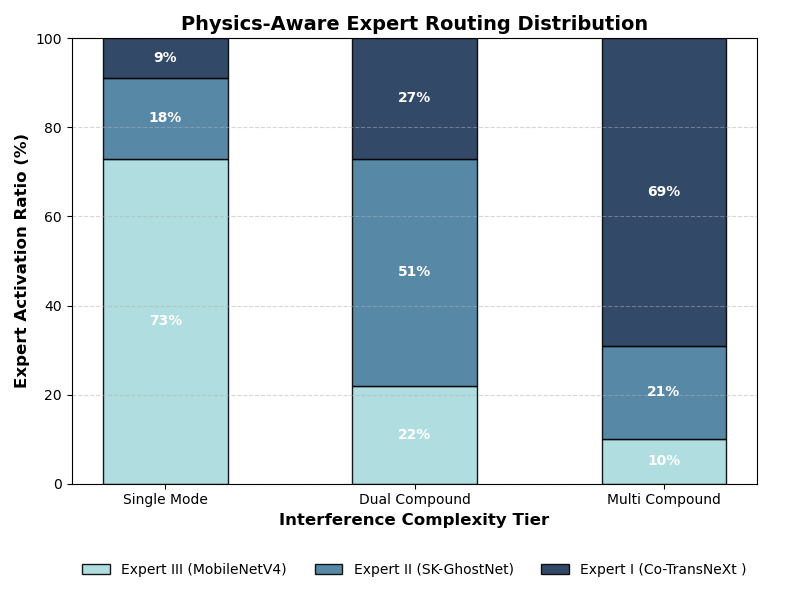}
    \caption{Physics-Guided Expert Routing Distribution. The gating network adaptively assigns lightweight experts (MobileNetV4) to simple jamming primitives and heavy-weight experts (TransNeXt) to complex compound signals, validating the cognitive resource allocation strategy.}
    \label{fig:router_dist}
\end{figure}

To validate the effectiveness and explainability of the proposed "Physics-Guided" gating mechanism, we analyze the expert activation distribution across different interference complexity tiers, as visualized in Fig. \ref{fig:router_dist}. The router demonstrates a strong physical correlation between the spectral entropy of the input and the computational capacity of the selected expert. Specifically, in the low-complexity regime (single-mode), the lightweight Expert III (MobileNetV4) processes the majority of samples (73\%), significantly conserving energy when high-capacity modeling is unnecessary. As complexity increases to the dual-compound regime, the activation shifts towards the mid-tier Expert II (SK-GhostNet), which handles 51\% of the samples, indicating that the router correctly identifies the need for multi-scale feature extraction without resorting to the heaviest expert. Conversely, facing saturated multi-compound interference, the router aggressively activates the heavy-weight Expert I (TransNeXt) for 69\% of the cases to leverage long-range attention for feature disentanglement. Crucially, this adaptive behavior aligns perfectly with human expert intuition: simple signals require minimal processing, while chaotic environments demand deep feature disentanglement. The router effectively learns this physical intuition without explicit supervision, confirming that PhyG-MoE utilizes intelligent, on-demand computing resources rather than random ensembling.

\section{Conclusion}
\label{sec:conclusion}

In this paper, we have proposed PhyG-MoE, a dynamic framework designed to balance recognition accuracy and computational efficiency for GNSS interference classification. Unlike traditional static models that process all inputs with the same computational cost, our approach uses a physics-guided gating mechanism based on PSD to allocate resources on demand. Simulation results on 21 jamming categories demonstrate that PhyG-MoE achieves an overall accuracy of 97.58\%. It exhibits exceptional robustness, particularly in identifying complex compound interference under low JNR conditions. Furthermore, the proposed dynamic strategy significantly reduces the average FLOPs compared to existing dual-stream networks, effectively minimizing the energy footprint. Our interpretability analysis confirms that the router operates with physical intuition, automatically assigning lightweight experts to simple signals and reserving heavy-weight computation for saturated ones. These characteristics make PhyG-MoE a practical and energy-efficient solution for resource-constrained GNSS receivers in the SAGIN.


\begin{thebibliography}{99}
\bibitem{saad2020vision} W. Saad, M. Bennis, and M. Chen, ``A Vision of 6G Wireless Systems: Applications, Trends, Technologies, and Open Research Problems,'' \textit{IEEE Network}, vol. 34, no. 3, pp. 134--142, May/June 2020.

\bibitem{nguyen20226g} D. C. Nguyen \textit{et al.}, ``6G Internet of Things: A Comprehensive Survey,'' \textit{IEEE Internet Things J.}, vol. 9, no. 1, pp. 359--383, Jan. 2022.

\bibitem{liu2024gnss} C. Liu, B. Ren, D. Fu, Y. Xie, and F. Chen, ``A GNSS Composite Interference Recognition Method Based on YOLOv5,'' in \textit{Proc. IEEE 6th Int. Conf. Civil Aviation Safety Inf. Technol. (ICCASIT)}, 2024, pp. 1157--1162.

\bibitem{nong2023adaptive} L. Nong, J. Peng, W. Zhang, J. Lin, H. Qiu, and J. Wang, ``Adaptive Multi-Hypergraph Convolutional Networks for 3D Object Classification,'' \textit{IEEE Trans. Multimedia}, vol. 25, pp. 4842--4854, 2023.

\bibitem{song2025tfunet} J. Song \textit{et al.}, ``GNSS Multiple interference mitigation with TF-Unet method based on high-resolution interference sensing,'' \textit{IEEE Sensors J.}, 2025.

\bibitem{wesson2018gnss} K. D. Wesson, J. N. Gross, T. E. Humphreys, and B. L. Evans, ``GNSS Signal Authentication Via Power and Distortion Monitoring,'' \textit{IEEE Trans. Aerosp. Electron. Syst.}, vol. 54, no. 2, pp. 739--754, Apr. 2018.

\bibitem{gamba2019performance} M. T. Gamba and E. Falletti, ``Performance Comparison of FLL Adaptive Notch Filters to Counter GNSS Jamming,'' in \textit{Proc. Int. Conf. Localization GNSS (ICL-GNSS)}, 2019, pp. 1--6.

\bibitem{gao2016protecting} G. X. Gao, M. Sgammini, M. Lu, and N. Kubo, ``Protecting GNSS Receivers From Jamming and Interference,'' \textit{Proc. IEEE}, vol. 104, no. 6, pp. 1327--1338, June 2016.

\bibitem{kato2019optimizing} N. Kato \textit{et al.}, ``Optimizing Space-Air-Ground Integrated Networks by Artificial Intelligence,'' \textit{IEEE Wireless Commun.}, vol. 26, no. 4, pp. 140--147, Aug. 2019.

\bibitem{liu2018sagin} J. Liu, Y. Shi, Z. M. Fadlullah, and N. Kato, ``Space-Air-Ground Integrated Network: A Survey,'' \textit{IEEE Commun. Surveys Tuts.}, vol. 20, no. 4, pp. 2714--2741, 2018.

\bibitem{mehr2025deep} I. E. Mehr and F. Dovis, ``A Deep Neural Network Approach for Classification of GNSS Interference and Jamming,'' \textit{IEEE Trans. Aerosp. Electron. Syst.}, vol. 61, no. 2, pp. 1660--1678, Apr. 2025.

\bibitem{sun2024frft} K. Sun, M. Elhajj, and W. Y. Ochieng, ``A GNSS Anti-Interference Method Based on Fractional Fourier Transform,'' \textit{IEEE Trans. Aerosp. Electron. Syst.}, vol. 60, no. 5, pp. 5636--5650, Oct. 2024.

\bibitem{li2025multipolarization} X. J. Li \textit{et al.}, ``A Multipolarization Interference Suppression Strategy of GNSS Antenna Array Based on Polarization Equalization,'' \textit{IEEE Trans. Aerosp. Electron. Syst.}, vol. 61, no. 5, pp. 12412--12426, Oct. 2025.

\bibitem{borio2014multistate} D. Borio, ``A Multi-State Notch Filter for GNSS Jamming Mitigation,'' in \textit{Proc. Int. Conf. Localization GNSS (ICL-GNSS)}, 2014, pp. 1--6.

\bibitem{sun2024novel} K. Sun, B. Yu, L. Xu, M. Elhajj, and W. Y. Ochieng, ``A Novel GNSS Anti-Interference Method Using Fractional Fourier Transform and Notch Filtering,'' \textit{IEEE Trans. Instrum. Meas.}, vol. 73, pp. 1--17, 2024.

\bibitem{qin2020assessment} W. Qin, M. T. Gamba, E. Falletti, and F. Dovis, ``An Assessment of Impact of Adaptive Notch Filters for Interference Removal on the Signal Processing Stages of a GNSS Receiver,'' \textit{IEEE Trans. Aerosp. Electron. Syst.}, vol. 56, no. 5, pp. 4067--4082, Oct. 2020.

\bibitem{alvarez2025chirp} X. Alvarez-Molina, G. Seco-Granados, and J. A. Lopez-Salcedo, ``Chirp Interference Mitigation in Snapshot GNSS Receivers Using the Time-Frequency Ridge-Assisted Fractional Fourier Transform,'' in \textit{Proc. Int. Conf. Localization GNSS (ICL-GNSS)}, 2025, pp. 1--6.

\bibitem{xiao2025compound} Z. Xiao \textit{et al.}, ``Compound Interference Recognition for GNSS via Integrating Time-Frequency With Power Spectrum Features,'' \textit{IEEE Trans. Aerosp. Electron. Syst.}, vol. 61, no. 5, pp. 13010--13024, Oct. 2025.

\bibitem{zhang2011effect} J. Zhang and E. S. Lohan, ``Effect and Mitigation of Narrowband Interference on Galileo E1 Signal Acquisition and Tracking Accuracy,'' in \textit{Proc. Int. Conf. Localization GNSS (ICL-GNSS)}, 2011, pp. 1--6.

\bibitem{chen2024cognitive} F. Chen, Q. Zhou, L. Huang, and C. Ren, ``GNSS Cognitive Interference Mitigation Method Based on Deep Learning,'' in \textit{Proc. 5th Int. Conf. Electron. Commun. Artif. Intell. (ICECAI)}, 2024, pp. 1--6.

\bibitem{wang2018detection} P. Wang, E. Cetin, A. G. Dempster, Y. Wang, and S. Wu, ``GNSS Interference Detection Using Statistical Analysis in the Time-Frequency Domain,'' \textit{IEEE Trans. Aerosp. Electron. Syst.}, vol. 54, no. 1, pp. 416--428, Feb. 2018.

\bibitem{chen2022fingerprint} X. Chen, D. He, X. Yan, W. Yu, and T.-K. Truong, ``GNSS Interference Type Recognition With Fingerprint Spectrum DNN Method,'' \textit{IEEE Trans. Aerosp. Electron. Syst.}, vol. 58, no. 5, pp. 4745--4760, Oct. 2022.

\bibitem{li2025dualGCN} Z. Li \textit{et al.}, ``GNSS Jamming Attacks Recognition Based on Dual GCN With Adaptive Weight Learning,'' \textit{IEEE Sensors J.}, vol. 25, no. 13, pp. 26152--26165, July 2025.

\bibitem{song2025sensing} J. Song \textit{et al.}, ``GNSS Multiple Interference Mitigation With TF-Unet Method Based on High-Resolution Interference Sensing,'' \textit{IEEE Sensors J.}, 2025.

\bibitem{jia2025lightweight} Q. Jia, L. Zhang, and R. Wu, ``Lightweight GNSS Interference Classifier and Its Solution Under Few-Shot Conditions,'' \textit{IEEE Trans. Aerosp. Electron. Syst.}, vol. 61, no. 6, pp. 17426--17444, Dec. 2025.

\bibitem{jia2025lowpower} Q. Jia, L. Zhang, and R. Wu, ``Low-Power Interference Identification Based on Convolutional Neural Networks,'' \textit{IEEE Trans. Instrum. Meas.}, vol. 74, pp. 1--15, 2025.

\bibitem{sun2025tsvae} P. Sun \textit{et al.}, ``Multiparameter Joint GNSS Spoofing Detection Based on TSVAE,'' \textit{IEEE Trans. Aerosp. Electron. Syst.}, vol. 61, no. 2, pp. 3373--3388, Apr. 2025.

\bibitem{vandermerwe2024optimal} J. R. van der Merwe, D. C. Franco, T. Feigl, and A. R{\"u}gamer, ``Optimal Machine Learning and Signal Processing Synergies for Low-Resource GNSS Interference Classification,'' \textit{IEEE Trans. Aerosp. Electron. Syst.}, vol. 60, no. 3, pp. 2705--2720, June 2024.

\bibitem{song2025power} J. Song \textit{et al.}, ``Power-Enhanced GNSS Interference Mitigation With Sensed Equivalent Bandwidth Based on ASF Algorithm,'' \textit{IEEE Sensors J.}, vol. 25, no. 2, pp. 2886--2896, Jan. 2025.

\bibitem{dasilva2023nmf} F. B. da Silva, E. Cetin, and W. A. Martins, ``Radio Frequency Interference Mitigation via Nonnegative Matrix Factorization for GNSS,'' \textit{IEEE Trans. Aerosp. Electron. Syst.}, vol. 59, no. 4, pp. 3493--3506, Aug. 2023.

\bibitem{garzia2021subband} F. Garzia \textit{et al.}, ``Sub-Band AGC-Based Interference Mitigation,'' in \textit{Proc. Int. Conf. Localization GNSS (ICL-GNSS)}, 2021, pp. 1--6.

\bibitem{wang2017statistical} P. Wang, E. Cetin, A. G. Dempster, Y. Wang, and S. Wu, ``Time Frequency and Statistical Inference Based Interference Detection Technique for GNSS Receivers,'' \textit{IEEE Trans. Aerosp. Electron. Syst.}, vol. 53, no. 6, pp. 2865--2876, Dec. 2017.

\bibitem{mehr2025gru} I. E. Mehr, G. Caputo, D. Salza, M. Fantino, and F. Dovis, ``Towards a Faster GNSS Interference Classification: a GRU-Based Approach using Spectrograms,'' in \textit{Proc. IEEE/ION Position, Location Navigat. Symp. (PLANS)}, 2025, pp. 372--381.

\bibitem{dovis2011wavelet} F. Dovis and L. Musumeci, ``Use of Wavelet Transforms for Interference Mitigation,'' in \textit{Proc. Int. Conf. Localization GNSS (ICL-GNSS)}, 2011, pp. 1--6.

\bibitem{luo2024zak} Y. Luo \textit{et al.}, ``Zak-Transform-Based Adaptive Interference Extraction Method for GNSS Interference Mitigation,'' \textit{IEEE Trans. Aerosp. Electron. Syst.}, vol. 60, no. 4, pp. 4784--4794, Aug. 2024.

\bibitem{ferre2019jammer} R. M. Ferre, A. de la Fuente, and E. S. Lohan, ``Jammer Classification in GNSS Bands Via Machine Learning Algorithms,'' \textit{Sensors}, vol. 19, no. 22, 2019.

\bibitem{woo2023convnext} S. Woo \textit{et al.}, ``ConvNeXt V2: Co-designing and Scaling ConvNets with Masked Autoencoders,'' in \textit{Proc. IEEE/CVF Conf. Comput. Vis. Pattern Recognit. (CVPR)}, 2023, pp. 16133--16142.

\bibitem{cai2019waterfall} Y. Cai \textit{et al.}, ``Jamming Pattern Recognition Using Spectrum Waterfall: A Deep Learning Method,'' in \textit{Proc. IEEE 5th Int. Conf. Comput. Commun. (ICCC)}, 2019, pp. 2113--2117.

\bibitem{han2020ghostnet} K. Han \textit{et al.}, ``GhostNet: More Features from Cheap Operations,'' in \textit{Proc. IEEE/CVF Conf. Comput. Vis. Pattern Recognit. (CVPR)}, 2020, pp. 1580--1589.

\bibitem{qin2024mobilenetv4} D. Qin \textit{et al.}, ``MobileNetV4: Universal Models for the Mobile Ecosystem,'' \textit{arXiv preprint arXiv:2404.10518}, 2024.

\bibitem{howard2019mobilenetv3} A. Howard \textit{et al.}, ``Searching for MobileNetV3,'' in \textit{Proc. IEEE/CVF Int. Conf. Comput. Vis. (ICCV)}, 2019, pp. 1314--1324.

\bibitem{an2025moe} T. An \textit{et al.}, ``An MoE-Driven Unified Image Restoration Framework for Adverse Weather Conditions,'' \textit{IEEE Trans. Circuits Syst. Video Technol.}, 2025.

\bibitem{xu2023slmoe} J. Xu \textit{et al.}, ``SL-MoE: A Two-Stage Mixture-of-Experts Sequence Learning Framework for Forecasting Rapid Intensification of Tropical Cyclone,'' in \textit{Proc. IEEE Int. Conf. Acoust., Speech, Signal Process. (ICASSP)}, 2023, pp. 1--5.

\bibitem{lepikhin2020gshard} D. Lepikhin \textit{et al.}, ``GShard: Scaling Giant Models with Conditional Computation and Automatic Sharding,'' \textit{arXiv preprint arXiv:2006.16668}, 2020.

\bibitem{liu2019deep} Q. Liu, ``Deep Learning and Recognition of Radar Jamming Based on CNN,'' in \textit{Proc. 12th Int. Symp. Comput. Intell. Design (ISCID)}, 2019, pp. 212--215.

\bibitem{li2019selective} X. Li, W. Wang, X. Hu, and J. Yang, ``Selective Kernel Networks,'' in \textit{Proc. IEEE/CVF Conf. Comput. Vis. Pattern Recognit. (CVPR)}, 2019, pp. 510--519.

\bibitem{li2023largesk} Y. Li \textit{et al.}, ``Large Selective Kernel Network for Remote Sensing Object Detection,'' in \textit{Proc. IEEE/CVF Int. Conf. Comput. Vis. (ICCV)}, 2023, pp. 1--10.

\bibitem{yang2023complexsk} H. Yang, Y. Zhang, T. Zhao, W. Zhang, and Z. He, ``Selective Kernel Fusion Complex-Valued CNN for Modulation Recognition,'' in \textit{Proc. IEEE 34th Annu. Int. Symp. Pers., Indoor Mobile Radio Commun. (PIMRC)}, 2023, pp. 1--6.

\bibitem{shi2024transnext} D. Shi, ``TransNeXt: Robust Foveal Visual Perception for Vision Transformers,'' \textit{arXiv preprint arXiv:2311.17132}, 2024.

\bibitem{yuan2021t2t} L. Yuan \textit{et al.}, ``Tokens-to-Token ViT: Training Vision Transformers from Scratch on ImageNet,'' in \textit{Proc. IEEE/CVF Int. Conf. Comput. Vis. (ICCV)}, 2021, pp. 558--567.

\bibitem{wang2026cluster} Z. Wang, G. Sun, Y. Wang, H. Yu, and D. Niyato, ``Cluster-Based Multi-Agent Task Scheduling for Space-Air-Ground Integrated Networks,'' \textit{IEEE Trans. Cogn. Commun. Netw.}, vol. 12, pp. 29--42, 2026.

\bibitem{liang2025generative} C. Liang \textit{et al.}, ``Generative AI-Driven Semantic Communication Networks: Architecture, Technologies, and Applications,'' \textit{IEEE Trans. Cogn. Commun. Netw.}, vol. 11, no. 1, pp. 27--42, Feb. 2025.

\bibitem{liu2026star} Y. Liu \textit{et al.}, ``STAR-RIS Enabled ISAC Systems With RSMA: Joint Rate Splitting and Beamforming Optimization,'' \textit{IEEE Trans. Cogn. Commun. Netw.}, vol. 12, pp. 312--324, 2026.

\bibitem{wang2026uav} C. Wang \textit{et al.}, ``UAV-Assisted Federated Learning With Robust Resource and Trajectory Optimization Under Location Uncertainties,'' \textit{IEEE Trans. Cogn. Commun. Netw.}, vol. 12, pp. 1282--1295, 2026.

\bibitem{zhang2026joint} Y. Zhang \textit{et al.}, ``Joint Deployment and Migration of Service Function Chains for Mobility-Aware Services in an Edge-Cloud Environment,'' \textit{IEEE Trans. Cogn. Commun. Netw.}, vol. 12, pp. 237--251, 2026.

\bibitem{tang2026digital} Y. Tang, K. Wang, D. Niyato, W. Chen, and G. K. Karagiannidis, ``Digital Twin-Assisted Federated Learning With Blockchain in Multi-Tier Computing Systems,'' \textit{IEEE Trans. Cogn. Commun. Netw.}, vol. 12, pp. 4517--4530, 2026.


\bibitem{11353414} Y. Xiu \textit{et al.}, ``Robust Optimization for Movable Antenna-Aided Cell-Free ISAC With Time Synchronization Errors,'' \textit{IEEE Trans. Wireless Commun.}, vol. 25, pp. 10082--10097, 2026.

\bibitem{11355857} Y. Xiu \textit{et al.}, ``Power Source Allocation for RIS-aided Integrating Sensing, Communication, and Power Transfer Communication Systems Based on NOMA,'' \textit{IEEE Trans. Mob. Comput.}, pp. 1--14, 2026.

\bibitem{11316665} Y. Xiu \textit{et al.}, ``Movable Antenna-Aided Cooperative ISAC Network with Time Synchronization Error and Imperfect CSI,'' \textit{IEEE Trans. Commun.}, p. 1, 2025.

\bibitem{11098592} Y. Xiu \textit{et al.}, ``Latency Minimization for Movable Relay-Aided D2D-MEC Communication Systems,'' \textit{IEEE Trans. Mob. Comput.}, vol. 25, no. 1, pp. 533--549, 2026.

\bibitem{11207524} Y. Xiu, W. Lyu, P. L. Yeoh, Y. Ai, and N. Wei, ``Secure Enhancement for RIS-Aided UAV with ISAC: Robust Design and Resource Allocation,'' \textit{IEEE Trans. Veh. Technol.}, pp. 1--16, 2025.

\bibitem{11346858} Z. Zhang, Y. Xiu, Z. Dong, J. Yin, M. J. Khabbaz, C. Assi, and N. Wei, ``Crosstalk-Resilient Beamforming for Movable Antenna Enabled Integrated Sensing and Communication,'' \textit{IEEE Wireless Commun. Lett.}, p. 1, 2026.

\bibitem{11316633} Z. Zhang, Y. Xiu, P. L. Yeoh, G. Liu, Z. Wu, and N. Wei, ``Distortion-Aware Hybrid Beamforming for Integrated Sensing and Communication,'' \textit{IEEE Commun. Lett.}, vol. 30, pp. 682--686, 2026.

\bibitem{11108293} X. Dong, W. Lyu, R. Yang, Y. Xiu, W. Mei, and Z. Zhang, ``Movable Antenna Enhanced Secure Simultaneous Wireless Information and Power Transfer,'' \textit{IEEE Commun. Lett.}, vol. 29, no. 10, pp. 2356--2360, 2025.

\bibitem{10797657} Y. Xiu \textit{et al.}, ``Robust Beamforming Design for Near-Field DMA-NOMA mmWave Communications With Imperfect Position Information,'' \textit{IEEE Trans. Wireless Commun.}, vol. 24, no. 2, pp. 1678--1692, 2025.

\bibitem{11220909} S. Yang \textit{et al.}, ``Flexible WMMSE Beamforming for MU-MIMO Movable Antenna Communications,'' \textit{IEEE Trans. Signal Process.}, vol. 73, pp. 4479--4491, 2025.
\end{thebibliography}

\end{document}